\documentclass[10pt,journal,compsoc]{IEEEtran}
\usepackage{amsfonts,amssymb,amsbsy,textcomp,marvosym,picins, amsmath,caption,threeparttable,amsthm,subfigure}
\usepackage{eurosym,mathrsfs,fancyhdr,CJK,multicol,graphics,indentfirst,color,bm,upgreek,booktabs,graphicx}
\usepackage{algorithm}
\usepackage{algorithmic}
\usepackage{subfigure}
\usepackage{todonotes}

\newtheorem{problem}{Problem}

\ifCLASSOPTIONcompsoc
  \usepackage[nocompress]{cite}
\else
  \usepackage{cite}
\fi
\ifCLASSINFOpdf
\else
\fi
\begin{document}
%
\title{P-ADMMiRNN: Training RNN with Stable Convergence via An Efficient and Paralleled ADMM Approach\thanks{Part of this article was published on "European Conference on Machine Learning and Principles and Practice of Knowledge Discovery in Databases", 2020\cite{tang2020admmirnn}}}
%
%
%
%

\author{Yu~Tang$^1$,
        Zhigang Kan$^1$,
        Dequan Sun$^1$,
        Jingjing Xiao,
        Zhiquan Lai$^1$,
        Linbo Qiao$^1\dagger$,
        and~Dongsheng Li$^1$
$1$ National University of Defense Technology, Changsha, China \\
$2$ Army Medical University (Third Military Medical University), Chongqing, China
\thanks{$\dagger$: corresponding authors. e-mail: qiao.linbo@nudt.edu.cn}
}

\IEEEtitleabstractindextext{%
\begin{abstract}
It is hard to train Recurrent Neural Network~(RNN) with stable convergence and avoid gradient \textit{vanishing} and \textit{exploding} problems, as the weights in the recurrent unit are repeated from iteration to iteration.
Moreover, RNN is sensitive to the initialization of weights and bias, which brings difficulties in training. 
The Alternating Direction Method of Multipliers~(ADMM) has become a promising algorithm to train neural networks beyond traditional stochastic gradient algorithms with the gradient-free features and immunity to unsatisfactory conditions.
However, ADMM could not be applied to train RNN directly since the state in the recurrent unit is repetitively updated over timesteps.
Therefore, this work builds a new framework named ADMMiRNN upon the unfolded form of RNN to address the above challenges simultaneously. We also provide novel update rules and theoretical convergence analysis. 
We explicitly specify essential update rules in the iterations of ADMMiRNN with constructed approximation techniques and solutions to each sub-problem instead of vanilla ADMM.
Numerical experiments are conducted on MNIST, IMDb, and text classification tasks. ADMMiRNN achieves convergent results and outperforms the compared baselines.
Furthermore, ADMMiRNN trains RNN more stably without gradient vanishing or exploding than stochastic gradient algorithms.
We also provide a distributed paralleled algorithm regarding ADMMiRNN, named P-ADMMiRNN, including Synchronous Parallel ADMMiRNN~(SP-ADMMiRNN) and Asynchronous Parallel ADMMiRNN~(AP-ADMMiRNN), which is the first to train RNN with ADMM in an asynchronous parallel manner. 
The source code is publicly available. 
\end{abstract}

\begin{IEEEkeywords}
ADMMiRNN, gradient \textit{vanishing} and \textit{exploding}, AP-ADMMiRNN, SP-ADMMiRNN

\end{IEEEkeywords}}

\maketitle

\IEEEdisplaynontitleabstractindextext

%
\IEEEpeerreviewmaketitle

\IEEEraisesectionheading{
\section{Introduction}\label{sec:introduction}}

%
%
%
%

\IEEEPARstart{R}{urrent} Neural Network~(RNN)~\cite{elman1990finding} has made great progress in various fields, namely language modelling, text classification~\cite{Lai2015TC}, event extraction~\cite{Nguyen2016JRNN}, and various real-world applications~\cite{graves2007multi,mikolov2010recurrent}. 
Although RNN models have been widely used, 
it is still difficult to train RNN models because of the \textit{vanishing gradients} and \textit{exploding gradients} problems\footnote{More information about \textit{vanishing gradients} and \textit{vanishing gradients} could be found in~\cite{bengio1994learning}}.
Moreover, RNN models are sensitive to the weights and biases~\cite{sutskever2013importance}, which may not converge with poor initialization.

Nowadays, gradient-based training algorithms are widely used in deep learning~\cite{lecun2015deep}, 
such as Stochastic Gradient Descent (SGD)~\cite{robbins1951stochastic}, Adam~\cite{Kingma2014Adam}, RMSProp~\cite{tieleman2012lecture}. 
However, 
they still suffer from \textit{vanishing} or \textit{exploding gradients}.
Compared with the traditional gradient-based optimization algorithms, the Alternating Direction Method of Multipliers~(ADMM) is a much more robust method to train deep neural networks.
It has been recognized as a promising method to alleviate \textit{vanishing gradients} and \textit{exploding gradients} problems and exert a tremendous fascination on researchers. Besides, ADMM is also immune to poor conditioning with gradient-free technique~\cite{taylor2016training}. Distributed ADMM is also proposed in recent years~\cite{boyd2011distributed,wei2013}. 
In addition, ADMM is a distributed-friendly algorithm and has drawn much attention from researchers. 
\cite{chen2018fully} develops a distributed approach for virtual power plant~(VPP) problems. 
\cite{li2019distributed} decomposes the regression problem into several subproblems and solves the communication latency and bandwidth cost in conventional cloud computing. 
As for neural networks, \cite{hosseini2014online} tests online ADMM on a network with linear constraints. This paper also extends distributed ADMM to a distributed setting and distributed gradient descent.  

In light of these properties of ADMM and to alleviate the problems mentioned above in RNN simultaneously, we are motivated to train RNN models with ADMM. However, it is not easy to apply ADMM to RNNs directly due to the recurrent state compared with MLP and CNN~\cite{krizhevsky2012imagenet}. 
The recurrent states are updated over timesteps instead of iterations, which is incompatible with ADMM.
Therefore, we propose a new framework named ADMMiRNN with theoretical analysis to tackle this problem. 

In distributed deep learning, data parallelism is an easy way to achieve but suffers from inter-GPU communication. The implementation of model parallelism needs replications of the activations in the neural networks, resulting in vast redundancy across GPUs. On the other hand, model parallelism is hard to achieve though it could reduce the communication cost because of the gradients and backpropagation~\cite{chen2018efficient}. In addition, we are also motivated to achieve model parallelism in RNN via ADMM and present Paralleled ADMMiRNN~(P-ADMMiRNN), including SP-ADMMiRNN and AP-ADMMiRNN, which are brought out to get a better training speed. Since there is much dependency between each parameter in ADMMiRNN, we choose to update parameters with the gradients in the last iteration in a parallel fashion and summarize it P-ADMMiRNN, which is inspired by \textit{Decoupled Parallel Backpropagation} (DDG)~\cite{huo2018decoupled}. 
Experimental comparisons between ADMMiRNN and some typical stochastic gradient algorithms, such as SGD and Adam, illustrate that ADMMiRNN avoids the \textit{vanishing gradients} and \textit{exploding gradients} problems and surpasses traditional stochastic gradient algorithms in terms of stability and efficiency. Besides, experiments evaluate the efficiency of P-ADMMiRNN, and comparisons among vanilla ADMMiRNN, SP-ADMMiRNN, and AP-ADMMiRNN show the convergence of our method.  

The main contributions of this work are summarized below:
\begin{itemize}
\item We propose a new framework named ADMMiRNN to train RNN models via ADMM.
ADMMiRNN is built upon the unfolded RNN unit, which is a remarkable feature of RNN, and could settle the problems of \textit{gradient vanishing} or \textit{exploding} and sensitive parameter initialization in RNN at the same time. Instead of using vanilla ADMM, some practical skills in our solution also help converge. In this way, the problem caused by the recurrent state is perfectly alleviated. 
To the best of our knowledge, we are the first to handle RNN training problems using ADMM, a gradient-free approach that brings significant advantages on stability beyond traditional stochastic gradient algorithms.
This gradient-free method also brings convenience into our paralleled work. 
\item The update rules of ADMMiRNN are presented, and also the theoretical analysis of convergence property is given. Our analysis ensures that ADMMiRNN achieves an efficient and stable result. Moreover, the framework proposed in this work could be applied to various RNN-based tasks.
\item We could train ADMMiRNN in a distributed paralleled manner and name it P-ADMMiRNN. Synchronous Parallel ADMMiRNN~(SP-ADMMiRNN) and Asynchronous Parallel ADMMiRNN~(AP-ADMMiRNN) are included. Unlike traditional distributed methods, P-ADMMiRNN is a DDG-like training method in which parameters are updated with the gradients in the last iteration. 
Moreover, P-ADMMiRNN is a new form of model parallelism, but its gradient-free property simplifies the training process and helps avoid the harmful effects of gradients and backpropagation in traditional model parallelism.
Experimental comparison between them evaluate that AP-ADMMiRNN converges faster than SP-ADMMiRNN. They both work better than vanilla ADMMiRNN. This is also the first systematic analysis of distributed training of ADMM in deep learning. 
\item Based on our theoretical analysis, numerical experiments are conducted on several real-world datasets. The experiment results demonstrate the efficiency and stability of the proposed ADMMiRNN beyond some other typical optimizers. Experimental results also verify our P-ADMMiRNN algorithms. 
\end{itemize}

\begin{figure*}[htbp]
\centering
\subfigure[]{
\includegraphics[width=0.15\textwidth]{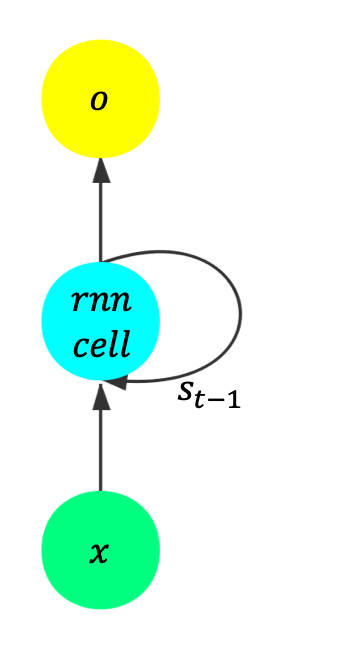}\label{fig:rnncell}
}
\quad
\quad \quad \quad \quad
\subfigure[]{
\includegraphics[width=0.27\textwidth]{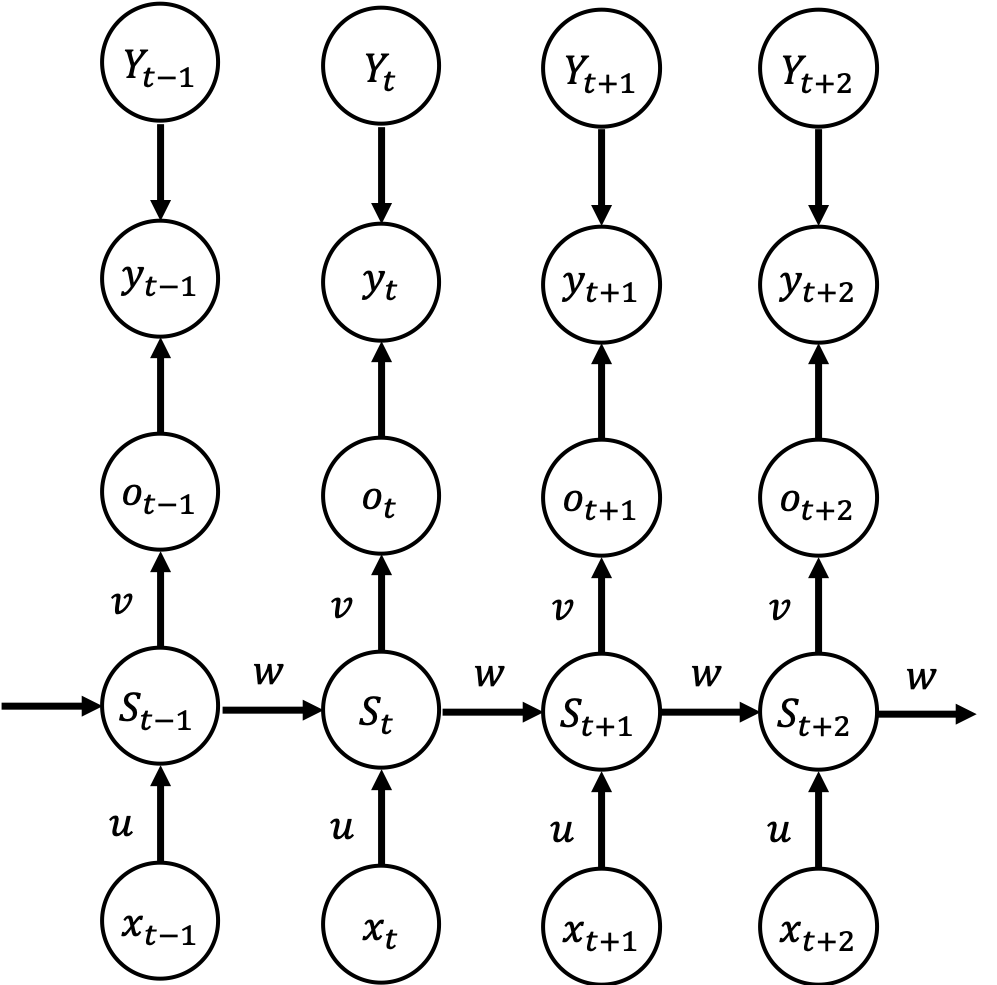}\label{fig:unfold}
}
\caption{Two different forms of RNN. \textbf{a}: The typical RNN cell. \textbf{b}: The unfolded form of Fig.~\ref{fig:rnncell}, which is functionally identical to the original form~\cite{goodfellow2016deep}.}
\end{figure*}

\section{Background and Related Work}\label{sec:related_work}

\subsection{Background}\label{sec:background}
The fundamental research about Recurrent Neural Networks was published in the 1980s.
RNNs are powerful to model problems with a defined order but no clear concept of time, with a variant of Long short-term memory(LSTM)~\cite{hochreiter1997long}. 
In~\cite{bengio1994learning}, they argued that it was difficult to train RNN models due to the \textit{vanishing gradients} and \textit{exploding gradients}.
Moreover, since RNN is sensitive to the initialization of weights and bias, those parameters should be initialized according to the input data~\cite{sutskever2013importance}. In \cite{pascanu2013difficulty}, they also state some difficulties in train RNNs. There is still a lack of a method to solve these above problems in RNN at the same time until now. 

In deep learning, optimization algorithms are commonly used to satisfy the performance of deep neural networks. Among them, stochastic gradient algorithms are mostly used in deep learning as the result of backpropagation. 
Stochastic gradient algorithms utilize the gradients of the loss function and update relative weights in each iteration. The gradients play an essential role in the training process. 
For example, Stochastic Gradient Descent~(SGD), one of the most typical optimization algorithms, gets the gradients of each iteration and applies them to updating the training weights.
Besides SGD, some two-order stochastic gradient algorithms are also commonly used in deep learning~\cite{Kingma2014Adam}. 
However, if the gradients are going to zero or infinity, it will incur an unconvergent training process, which is known as \textit{gradient vanishing} or \textit{gradient exploding} problems, which is inevitable in these gradient-based optimization algorithms.  

ADMM was first introduced in~\cite{gabay1976dual}.
Its convergence was established in~\cite{gabay1983augmented,glowinski1989augmented}.
Since ADMM can decompose large problems with constraints into several small ones, it has been one of the most powerful optimization frameworks. It shows a multitude of well-performed properties in plenty of fields, such as machine learning~\cite{boyd2011distributed}, signal processing~\cite{sun2018iteratively} and tensor decomposition~\cite{goldfarb2014robust} and modal decomposition~\cite{masuyama2018modal}. 

In general, ADMM seeks to tackle the following problem:
\begin{equation}\label{general ADMM}
\min\limits_{x,y} f(x) + g(y),     \quad {\rm s.t.} \quad Ax+By = c.
\end{equation}
Here, $f: \mathbb{R}^{n_1} \rightarrow \mathbb{R} $ and $g: \mathbb{R}^{n_2} \rightarrow \mathbb{R}$ are usually assumed to be convex functions.
In Eq.~\eqref{general ADMM}, $A \in \mathbb{R}^{m \times n_1}$, $B \in \mathbb{R}^{m \times n_2}$, $c \in \mathbb{R}^m$, and $Ax+By = c$ is a linear constraint and $n_1$, $n_2$ are the dimensions of $x$, $y$ respectively.
It is solved by the Augmented Lagrangian Method which is formalized as: 
\begin{equation}\label{equ:into}
\begin{aligned}
    \mathcal{L}_\beta(x, y, \lambda) = & f(x) + g(y) + <\lambda, Ax+By-c>     \\
                                     & + \frac{\beta}{2}\|Ax+By-c\|^2,
\end{aligned}
\end{equation}
where $\beta$ is the penalty term, $\lambda$ is the Lagrangian multiplier.

\begin{table}[!t]
\caption{Important notations and corresponding descriptions.}
\label{tab:notations}
\centering
\begin{tabular*}{\linewidth}{c c}\hline\hline\hline
Notations    & Descriptions  \\
\hline
$t$         & the timestep         \\
$x_t$       & the input of RNN cells    \\
$o_t$       & the output of RNN cells    \\
$s_t$        & the state at timestep $t$    \\
$u$            & the weight corresponding to the input     \\
$w$            & the weight corresponding to the state     \\
$y_t$        & the prediction at timestep $t$   \\
$N$            & the cell numbers after unfolding   \\
$R$            & the loss function     \\
$\Omega(w)$    & the regularization term    \\
$\theta$    & $\{u,w,b,a,s,v,c,o\}$ \\
$k$         & the iteration count       \\
\hline\hline\hline
\end{tabular*}
\end{table}

\subsection{Related Work}

The Alternating Direction Method of Multipliers~(ADMM) is one of the optimization algorithms in machine learning. ADMM has shown its great power in plenty of fields of machine learning, including convex optimization~\cite{sun2018alternating} and nonconvex optimization~\cite{guan2018an}. 

Since ADMM was first proposed, plenty of theoretical and practical works have been developed in recent years~\cite{monteiro2010iteration}.
In 2016,~\cite{taylor2016training} proposed a new method to train neural networks using ADMM. They abandoned traditional optimizers and adopted ADMM, which trains neural networks in a robust and paralleled fashion. 
Furthermore,  ADMM was applied to deep learning and obtained a remarkable result~\cite{wang2019admm}. They provided a gradient-free method to train neural networks, gaining convergent and excellent performance. Their works prove that ADMM is a powerful optimization method for neural networks because of its gradient-free property. However, RNNs are not as simple as a Multilayer Perceptron. The recurrent state brings many challenges when solving RNN with ADMM. 

There have been few works concerning ADMM in deep neural networks in recent years.  \cite{taylor2016training} is the first to apply ADMM into linear neural networks and brings out a new training algorithm without gradients.~\cite{wang2019admm} gets much more improved and training deep neural networks with a fast and efficient ADMM algorithm named dlADMM. They are the first to apply ADMM to deep learning and propose \textit{backward-forward} updating rules and gets a speedup of convergence.
However, they only consider linear neural networks. RNN models are more complicated and require more careful analysis. 

Besides, since it is much simple to achieve ADMM in parallel~\cite{boyd2011distributed}, there is also much work related to the application of ADMM in parallel. In 2016, Chang et al. proposed Asynchronous Distributed ADMM~(AD-ADMM) for large-scale optimization, including its algorithms and convergence analysis~\cite{chang2016asynchronous}. They also provided a linear convergence analysis for large-scale optimization in~\cite{chang2016asynchronous2}. In \cite{jiang2019}, these authors used a dynamic scheduling strategy in the asynchronous ADMM algorithm for distributed optimization and this strategy improved the convergence speed and communication efficiency of ADMM in large-scale clusters. Compared with traditional data parallelism and model parallelism, ADMM in parallel is not only simple to implement but also does not suffer from the massive communication in data parallelism. 

Since there is not any work about training RNN with ADMM, not to mention training in a paralleled way. Therefore, we are also motivated to achieve training RNN with paralleled ADMM algorithms.

\section{ADMM for RNNs}\label{sec:ADMM_RNN}
\subsection{Notation}\label{subsec:notions}
Before we dive into the ADMM methods for RNNs, we establish notations in this work.
Considering a simple RNN cell as shown in Fig.~\ref{fig:rnncell}, at timestep $t>=1$, $x_t$ is the input of the RNN cell and $o_t$ is the related output, RNNs could be expressed as:
\begin{equation}
\begin{array}{cl}
&\Phi(x_1, x_2, \cdots, x_N, u, v, w, b, c) \\
=& vf(ux_N+ w(vf(ux_{N-1}+ w(\cdots vf(ux_0 \\
& +b) + c \cdots)+b) + c)+b) + c\\
&+\Omega(W),
\end{array}
\end{equation} 
where $f(\cdot)$ is an activation function and $v, u, w, b$ and ${c}$ are trainable parameters, $s_0=0$. These parameters are also unified. The recurrent state in RNNs varies over timesteps as well as iterations, which brings difficulties to applying ADMM into RNNs directly. We adopt an unfolding form of the RNN unit shown in Fig.~\ref{fig:unfold} and decouple these above parameters into three sub-problems.
Normally, at timestep $t$, the updates are listed in the following:
\begin{equation}
\begin{aligned}
    a_t & = ux_t + ws_{t-1} + b ,     \\
    s_t & = f(a_t)  ,                \\
    o_t & = vs_t + c ,               \\
\end{aligned}
\end{equation}
where $f(\cdot)$ is the activation function, such as ReLU~\cite{nair2010rectified}
or \textit{tanh}, usually \textit{tanh} in RNNs.
Necessary notations are summarized in Table~\ref{tab:notations}. 

In this paper, we consider RNN in an unfolding form and present a theoretical analysis based on it. 

For the sake of convenience, we define $\theta=\{u,w,b,a,s,v,c,o\}$ in the sequel. 
In term of applying ADMM into RNNs, assuming the RNN cell is unfolded into $N$ continuous cells, we try to solve the mathematical problem as follows:
\begin{problem}\label{problem:1}
\begin{equation}\label{equ:problem1}
\begin{aligned}
    & \min\limits_{\theta_t} \Phi(\theta_t) \equiv R(\theta_t) + \Omega(w) ,   \\
    & {\rm{s.t.}} \quad a_t = ux_t + ws_{t-1} + b, s_t = f(a_t), o_t = vs_t + c.
\end{aligned}
\end{equation}
\end{problem}
In Problem~\ref{problem:1}, $R(\theta_t)$ is the loss function which is convex and continuous, $\Omega(w)$ is the regularization term on the parameter $w$.
It is also a convex and continuous function.
Rather than solving Problem~\ref{problem:1} directly, we can relax it by adding an $l_2$ penalty term and transform Eq.~\eqref{equ:problem1} into 
\begin{problem}\label{problem:2}
\begin{equation}\label{equ:problem2}
    \begin{aligned}
        \min\limits_{\theta_t} & R(\theta_t) + \Omega(w) + \frac{\nu}{2}\sum_{t=1}^{N-1}(\|a_t-ux_t-ws_{t-1}-b\|^2   \\
        & + \|s_t-f(a_t)\|^2 + \|o_t-vs_t-c\|^2)  \\
    \end{aligned}
\end{equation}
\begin{equation}\nonumber
    \begin{aligned}
        {\rm{s.t.}} \quad a_N = ux_N + ws_{N-1} + b, s_N = f(a_N), o_N = vs_N + c,
    \end{aligned}    
\end{equation}
\end{problem}
where $\nu$ is a tuning parameter.
Compared with Problem~\ref{problem:1}, Problem~\ref{problem:2} is much easier to solve.
According to~\cite{wang2019admm}, the solution of Problem~\ref{problem:2} tends to be the solution of Problem~\ref{problem:1} when $\nu \rightarrow \infty$.
For simplicity and clarity, we often use $<\cdot,\cdot>$ to denote the inner product and $\Tilde{k} = k+1$.
For a positive semidefinite matrix $G$, we define the $G-$norm of a vector as $\|x\|_G=\|G^{1/2}x\|_2=\sqrt{x^TGx}$.
\subsection{ADMM Solver for RNN}\label{subsec:admm solver}
As aforementioned in Section~\ref{sec:related_work}, we explain that ADMM utilizes the Augmented Lagrangian Method to solve problems like Eq.~\eqref{equ:into}.
Similarly, we adopt the same way and present the corresponding Lagrangian function of Eq.~\eqref{equ:problem2}, namely Eq.~\eqref{equ:lagrangian}:
\begin{equation}\label{equ:lagrangian}
    \mathcal{L}_{\rho_1,\rho_2,\rho_3}(\theta)=R(o) + \Omega(w)+\phi(\theta_t),
\end{equation}
where $\phi(\theta_t)$ is defined in Eq.~\eqref{equ:phi}.
\begin{equation}\label{equ:phi}
    \begin{aligned}
        \phi(\theta_t) =&\frac{\nu}{2}\sum_{t=1}^{N-1}(\|a_t-ux_t-ws_{t-1}-b\|^2 + \|s_t-f(a_t)\|^2  \\   
        &  +\|o_t-vs_t-c\|^2) + <\lambda_1, a_N - ux_N - ws_{N-1}   \\
        &   - b> + <\lambda_2, s_N - f(a_N)> + <\lambda_3, o_N- \\ 
        & vs_N - c> + \frac{\rho_1}{2}\| a_N - ux_N - ws_{N-1} - b\|^2 +   \\
        &  \frac{\rho_2}{2}\|s_N - f(a_N)\|^2 + \frac{\rho_3}{2}\|o_N - vs_N -c\|^2.
    \end{aligned}
\end{equation}
Problem~\ref{problem:2} is separated into eight subproblems and could be solved through the updates of these parameters in $\theta_t$. Note that $u,w,b,v,c$ in $\theta$ are not changed over timestep $t$.
\begin{algorithm}[!t]
\caption{The training algorithm for ADMMiRNN.}
\label{alg:RNN algorithm}
\textbf{Input}: iteration $K$, input $x$, timestep $N$. \\
\textbf{Parameter}: $u$, $w$, $b$, $v$, $c$, $s_0$, $\lambda_1$, $\lambda_2$, and $\lambda_3$\\
\textbf{Output}: $u$, $w$, $b$, $v$, $c$,
\begin{algorithmic}[1] 
\STATE Initialize $k=0$, $u$, $w$, $b$, $v$, $c$, $s_0$, $\lambda_1$, $\lambda_2$, and $\lambda_3$.
\FOR{$k=1,2,\cdots, K$}
\FOR{$t=1,2,\cdots, N$}
\IF{$t<N$} 
    \STATE Update $o_t^{\Tilde{k}}$ in Eq.~\eqref{equ:update_ot}.
\ELSIF{$t=N$}
    \STATE Update $o_N^{\Tilde{k}}$ in Eq.~\eqref{equ:update_oN}.
\ENDIF
    \STATE Update $c^{\Tilde{k}}$ in Eq.~\eqref{equ:update_ct}.
    \STATE Update $v^{\Tilde{k}}$ in Eq.~\eqref{equ:update_vt}.
\IF{$t<N$} 
    \STATE Update $s_t^{\Tilde{k}}$ in Eq.~\eqref{equ:update_st}.
    \STATE Update $a_t^{\Tilde{k}}$ in Eq.~\eqref{equ:update_at}.
\ELSIF{$t=N$}
    \STATE Update $s_N^{\Tilde{k}}$ in Eq.~\eqref{equ:update_sN}.
    \STATE Update $a_N^{\Tilde{k}}$ in Eq.~\eqref{equ:update_aN}.
\ENDIF
    \STATE Update $b^{\Tilde{k}}$ in Eq.~\eqref{equ:update_bt}.
    \STATE Update $w^{\Tilde{k}}$ in Eq.~\eqref{equ:update_w}.
    \STATE Update $u^{\Tilde{k}}$ in Eq.~\eqref{equ:update_ut}.
    \STATE Update $u^{\Tilde{k}}$ in Eq.~\eqref{equ:update_ut}.
    \STATE Update $w^{\Tilde{k}}$ in Eq.~\eqref{equ:update_w}.
    \STATE Update $b^{\Tilde{k}}$ in Eq.~\eqref{equ:update_bt}.
\IF{$t<N$} 
    \STATE Update $a_t^{\Tilde{k}}$ in Eq.~\eqref{equ:update_at}.
    \STATE Update $s_t^{\Tilde{k}}$ in Eq.~\eqref{equ:update_st}.
\ELSIF{$t=N$}
    \STATE Update $a_N^{\Tilde{k}}$ in Eq.~\eqref{equ:update_aN}.
    \STATE Update $s_N^{\Tilde{k}}$ in Eq.~\eqref{equ:update_sN}.
\ENDIF    
    \STATE Update $v^{\Tilde{k}}$ in Eq.~\eqref{equ:update_vt}.
    \STATE Update $c^{\Tilde{k}}$ in Eq.~\eqref{equ:update_ct}.
\IF{$t<N$}    
    \STATE Update $o_t^{\Tilde{k}}$ in Eq.~\eqref{equ:update_ot}.
\ELSIF{$t=N$}
    \STATE Update $o_N^{\Tilde{k}}$ in Eq.~\eqref{equ:update_oN}.
\ENDIF
\ENDFOR
\STATE Update $\lambda_1^{\Tilde{k}}$ in Eq.~\eqref{equ:update_lambda1}.
\STATE Update $\lambda_2^{\Tilde{k}}$ in Eq.~\eqref{equ:update_lambda2}.
\STATE Update $\lambda_3^{\Tilde{k}}$ in Eq.~\eqref{equ:update_lambda3}.
\ENDFOR
\STATE \RETURN $u$, $w$, $b$, $v$, $c$,
\end{algorithmic}
\end{algorithm}
\begin{equation}\label{equ:g-norm_u}
\begin{aligned}
    u^{\Tilde{k}} \leftarrow & \arg\min\limits_{u} \frac{\nu}{2}\sum_{t=1}^{N-1}\|a_t-ux_t-ws_{t-1}-b\|^2+\frac{\rho_1}{2}\|a_N \\
    & -ux_N-ws_{N-1}-b-\lambda_1/\rho_1\|^2+\frac{N}{2}\|u-u^k\|_{\textbf{G}}^2.   \\
\end{aligned}
\end{equation}
It is equivalent to the linearized proximal point method inspired by~\cite{rockafellar1976monotone}:
\begin{equation}\label{equ:update_ut}
    \begin{aligned}
        u^{\Tilde{k}} \leftarrow & \arg\min\limits_{u} \frac{Nr}{2}\|u-u^k\|^2+\nu(u-u^k)^T\sum_{t=1}^{N-1}[(x_t^k)^T   \\
        & (a_t-u^kx_t^k-w^ks_{t-1}^k-b^k)]+\rho_1(u-u^k)^T[(x_N^k)^T   \\
        & (a_N^k-u^kx_N^k-w^ks_{N-1}^k-b^k-\lambda_1^k/\rho_1)].  \\
    \end{aligned}
\end{equation}

Consequently, these parameters are supposed to update over iterations. To make it clear, we only describe the specific update rules for $u, a$ and $s$ in the following subsections because there are some valuable and typical skills in these subproblems, and analysis of the other parameters detailed in Appendix A in the supplementary materials is similar.

\subsubsection{Update $u$}\label{sec:update_u}
We begin with the update of $u$ in Eq.~\eqref{equ:lagrangian} at iteration $k$.
In Eq.~\eqref{equ:phi}, $u$ and $x_t$ are coupled.
As a result, we need to calculate the pseudo-inverse of the (rectangular) matrix $x_t$, making it harder for the training process.
In order to solve this problem, we define $\textbf{G}=rI_{d}-\rho_1 x_{t}^Tx_{t}$ and replace it with Eq.~\eqref{equ:g-norm_u}. 
In this way, the update of $u$ significantly speeds up than the vanilla ADMM. It is worth noting that $r$ needs to be appropriately set, and $r$ could also affect the performance of ADMMiRNN. 

\subsubsection{Update $a$}

Adding a proximal term similar to that in Section~\ref{sec:update_u}, if $t<N$, this could be done by
\begin{equation}\label{equ:update_at}
    \begin{aligned}
        a_t^{\Tilde{k}} \leftarrow & \arg\min\limits_{a_t} \frac{r}{2}\|a_t-a_t^k\|^2+\nu (a_t-a_t^k)^T(a_t^k-u^kx_t^k \\
        & -w^ks_{t-1}^k-b^k)+\frac{\nu}{2}\|s_t-f(a_t)\|^2.
    \end{aligned}
\end{equation}
When $t=N$, 
\begin{equation}\label{equ:update_aN}
    \begin{aligned}
        a_N^{\Tilde{k}} \leftarrow & \arg\min\limits_{a_N} \frac{r}{2}\|a_N-a_N^k\|^2+\rho_{1} (a_N-a_N^k)^T(a_N^k-u^kx_N^k  \\
        & -w^ks_{N-1}^k-b^k-\lambda_1/\rho_1)+\frac{\rho_{2}}{2}\|s_N-f(a_N)-\lambda_2/\rho_2\|^2. \\
    \end{aligned}
\end{equation}
Here is a \textbf{trick}:
If $a_t$ is small enough, we have $f(a_t)=a_t$ as a result of the property of \textit{tanh} function. In this way, we could simplify the calculation of $a_t$.

\subsubsection{Update $s$}
The parameter $s$ represents the hidden state in the RNN cell shown in Fig.\ref{fig:rnncell}.
With regard to the update of $s$, there are $s_{t-1}$ and $s_t$ in Eq.~\eqref{equ:phi}.
However, we only consider $s_t$ in the RNN model. 
It is because $s_{t-1}$ has been updated in the last unit and would cause calculation redundancy in the updating process. This is another \textbf{trick} in our solution. 
Besides, $s_t$ also needs to be decoupled with $w$. 

If $t<N$, we could update $s_t$ through
\begin{equation}\label{equ:update_st}
    \begin{aligned}
        s_t^{\Tilde{k}} \leftarrow & \arg\min\limits_{s_t} \frac{r}{2}\|s_t-s_t^k\|^2+\nu(s_t-s_t^k)^T[(v^k)^T(o_t^k-v^ks_t^k\\
        & -c^k)]+\frac{\nu}{2}\|s_t-f(a_t)\|^2.   \\
    \end{aligned}    
\end{equation}
And when $t=N$, 
\begin{equation}\label{equ:update_sN}
    \begin{aligned}
        s_N^{\Tilde{k}} \leftarrow & \arg\min\limits_{s_N} \frac{r}{2}\|s_N-s_N^k\|^2+\rho_3(s_N-s_N^k)^T[(v^k)^T(o_N^k-\\
        & v^ks_N^k-c^k-\lambda_3^k/\rho_3)].   \\
    \end{aligned}    
\end{equation}

\subsubsection{Update Lagrangian Multipliers}
Similar to the parameters update, $\lambda_1$, $\lambda_2$ and $\lambda_3$ are updated as follows respectively:
\begin{subequations}
\begin{equation}\label{equ:update_lambda1}
    \lambda_1^{\Tilde{k}} = \lambda_1^k + \rho_1(a_N-ux_N-ws_{N-1}-b),
\end{equation}
\begin{equation}\label{equ:update_lambda2}
    \lambda_2^{\Tilde{k}} = \lambda_2^k + \rho_2(s_N-f(a_N)),
\end{equation}
\begin{equation}\label{equ:update_lambda3}
    \lambda_3^{\Tilde{k}} = \lambda_3^k + \rho_3(o_N-vs_N-c).
\end{equation}
\end{subequations}

\begin{figure*}[tbp]
\centering
\subfigure[SP-ADMMiRNN.]{
\includegraphics[width=0.45\textwidth]{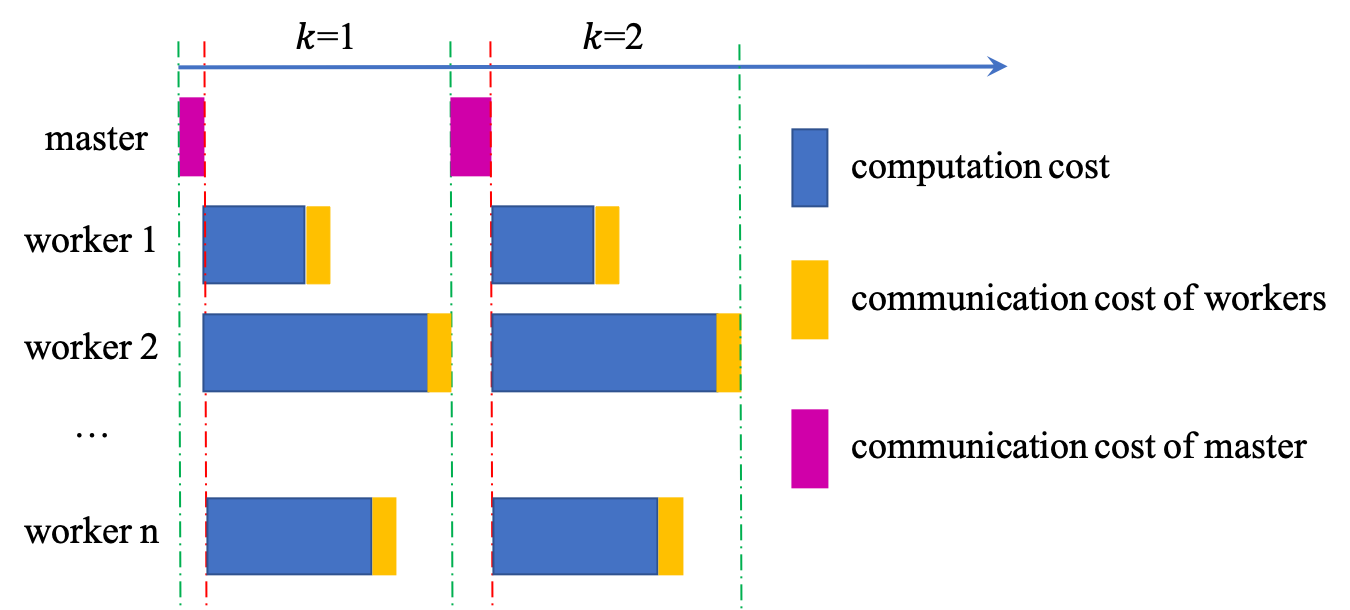}\label{fig:sync parallel}
}
\subfigure[AP-ADMMiRNN.]{
\includegraphics[width=0.45\textwidth]{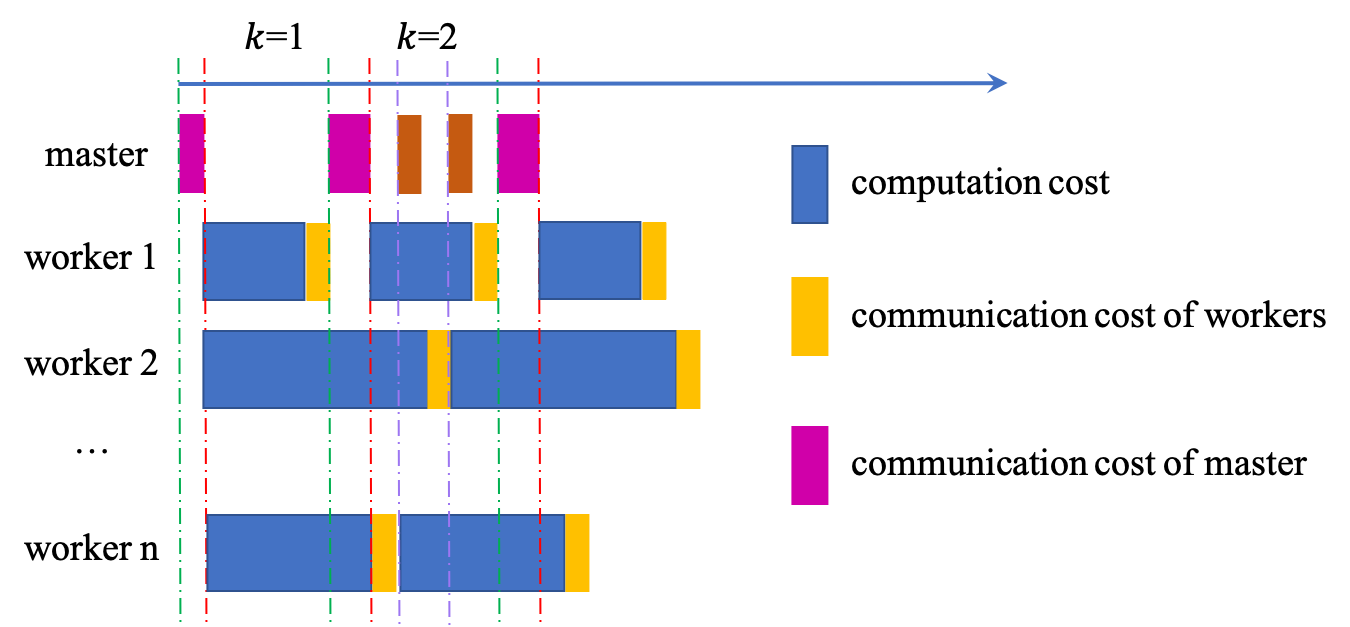}\label{fig:async parallel}
}
\caption{Illustration of Synchronous and Asynchronous Paralleled ADMMiRNN. It is much more computing efficient to train RNN in an asyncronous manner.}\label{fig:sync and async}
\end{figure*}

\subsubsection{Algorithm}
Generally, we update the above parameters in two steps.
First, these parameters are update in a backward way, namely $o\rightarrow c \rightarrow v \rightarrow s \rightarrow a \rightarrow b \rightarrow w \rightarrow u$.
Afterwards, ADMMiRNN reverses the update direction in $u\rightarrow w \rightarrow b \rightarrow a \rightarrow s \rightarrow v \rightarrow c \rightarrow o$.
After all those variables in an RNN cell update, the Lagrangian multipliers update. Proceeding with the above steps, we could arrive at the algorithms for ADMMiRNN, which is outlined in Algorithm~\ref{alg:RNN algorithm}.

\subsection{P-ADMMiRNN}\label{sec:paralleled algorithms}

This section introduces the paralleled distributed training algorithms in ADMMiRNN, including Asynchronous Parallel ADMMiRNN~(AP-ADMMiRNN) and Synchronous Parallel ADMMiRNN~(SP-ADMMiRNN). 
Data parallelism and model parallelism have their own pros and cons. Data parallelism splits the training data into several subsets and is much simpler to implement than model parallelism. However, it consumes much memory and suffers from communication overheads. The communication between workers in model parallelism is significantly less than data parallelism, but traditional model parallelism suffers a lot from the backpropagation and faces the staleness issue, resulting in the instability~\cite{chen2018efficient}. 
In our distributed algorithms, we adopt a ``\textit{master-worker}'' method. The \textit{master} is responsible for the initialization of parameters and managing those parameters while the \textit{worker} does all the calculations in ADMMiRNN. So the algorithm of the \textit{master} in AP-ADMMiRNN and SP-ADMMiRNN could both be presented in Algorithm~\ref{alg:master}.  
\begin{algorithm}[H]
\caption{The master algorithm in AP-ADMMiRNN and SP-ADMMiRNN.} \label{alg:master}
\hspace*{0.02in} 
\textbf{Input:} the iteration number $K$, input data $x$, send queue $Q_s$, receive queue $Q_r$. \\
\hspace*{0.02in}   
\textbf{Output:} $loss$ and $accuracy$.\\
\begin{algorithmic}[1]
\STATE Init: $u, w, s_0, b, v, c$.
\STATE $Q_s$ sends all the hyperparameters to workers.
	\STATE $Q_s$ sends $u, w, s_0 b, v, c$ to workers.
\FOR{t = 1 to $K$} 
	\STATE $Q_r$ receives $u, w, s_{t}, b, v, c$ from workers.
	\STATE Compute $loss$ and $accuracy$.
\ENDFOR
\RETURN $loss$ and $accuracy$.
\end{algorithmic}
\end{algorithm}
\begin{algorithm}[H]
\caption{The worker algorithms in SP-ADMMiRNN.} \label{alg:sp-admmirnn worker}
\hspace*{0.02in} 
\textbf{Input:} the worker number $N_{sp}$, send queue $Q_s$, receive queue $Q_r$. \\
\hspace*{0.02in}  
\textbf{Output:} $u, w, s_{t}, b, v, c$.  \\
\begin{algorithmic}[1]
\STATE $Q_r$ receives hyperparameters from the master.  
\STATE $Q_r$ receives $u, w, s_{t}, b, v, c$ from the master.
\STATE Assign all the parameters to $N_{sp}$ workers. 
\FOR{t = 1 to $K$} 
    \FOR{i = 1 to $N_{sp}$}
        \STATE worker $N_i$ updates its parameters.
    \ENDFOR
	\STATE $Q_s$ sends $u, w, s_{t}, b, v, c$ to master.
\ENDFOR
\end{algorithmic}
\end{algorithm}

\subsubsection{SP-ADMMiRNN}
\begin{figure}
    \centering
    \includegraphics[width=0.4\textwidth]{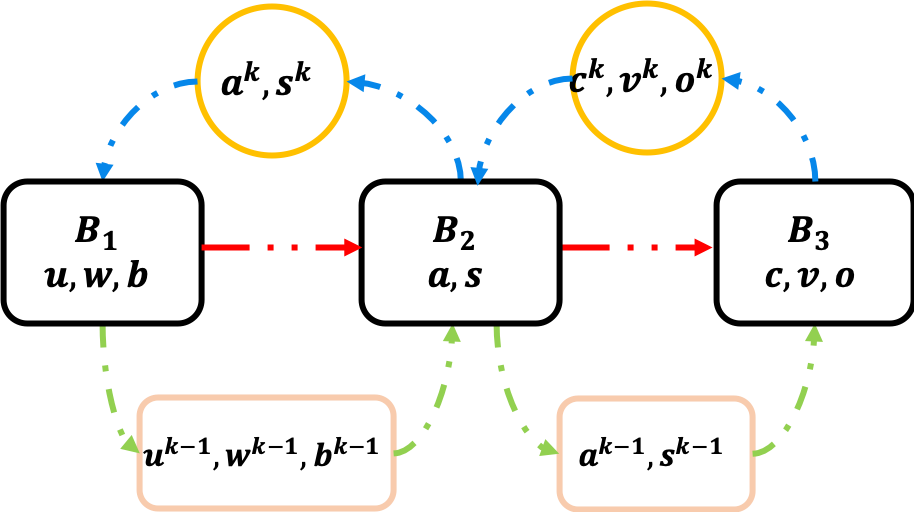}
    \caption{An example of Synchronous Parallel ADMMiRNN. There are three workers in this example, $B_1,B_2$, and $B_3$. These three workers are responsible for different parameter updates. At iteration $k\le 1$, $B_1$ sends $u^{k-1},w^{k-1}$ and $b^{k-1}$ to $B_2$. At the mean time, $B_2$ sends $a^{k-1}$ and $s^{k-1}$ to $B_3$. $B_1,B_2$, and $B_3$ update their parameters at the same time. After the updates, the relative parameters will be sent to their workers to continue the process.}
    \label{fig:sp example}
\end{figure}

Synchronous Parallel ADMMiRNN is a little different form traditional synchronous algorithms.
In our Synchronous Parallel ADMMiRNN, we split the training process into $n$ blocks, namely $B_1,B_2, \cdots, B_n$. These blocks could also be regarded as workers in the distributed system. All the workers calculate parts of updates of those parameters at the same time. At iteration $k$, The worker $B_m$ sends the parameters in last iteration to $B_{m+1}$, where $m=1,2, \cdots, n$. The parameters won't update until all of these parameters are computed. After all of these parameters are updated, they will be sent the former block so as to continue the training process. This is inspired by \cite{huo2018decoupled}. To make it clear, we describe an example in Fig.~\ref{fig:sp example}. 

The synchronous worker algorithms are summarized in Algorithm~\ref{alg:sp-admmirnn worker}. At first, the worker receives all the hyperparameters and parameters from the master. And then, these parameters would be assigned to $N_{sp}$ workers. After updating those parameters, they would be sent to master through the send queue $Q_s$ in worker.

\subsubsection{AP-ADMMiRNN}
SP-ADMMiRNN gets improved compared with vanilla ADMMiRNN. However, it is still limited by the slowest workers, especially when the workers have different computation and communication delays. Therefore, we are motivated to achieve Asynchronous Parallel ADMMiRNN~(AP-ADMMiRNN), and its worker algorithms are presented in Algorithm~\ref{alg:ap-admmirnn worker}. Algorithm~\ref{alg:ap-admmirnn worker} is slightly different from Algorithm~\ref{alg:sp-admmirnn worker} in Line 7. The send queue $Q_s$ in AP-ADMMiRNN would send those parameters, whichever has been updated. In this way, the convergence could speed up a lot. 

Fig.\ref{fig:sync and async} shows the detailed process of SP-ADMMiRNN and AP-ADMMiRNN. In SP-ADMMiRNN, the training time depends on the longest computation time. The other workers need to wait for the most time-consuming worker. It is clear that in SP-ADMMiRNN, the master takes much time waiting for the parameters in each iteration. However, in AP-ADMMiRNN, the parameters could be sent to the master as long as they are updated, saving much unnecessary time. The master is much busier than that in SP-ADMMiRNN.

\begin{algorithm}[H]
\caption{The worker algorithms in AP-ADMMiRNN.} \label{alg:ap-admmirnn worker}
\hspace*{0.02in} 
\textbf{Input:} the worker number $N_{ap}$, send queue $Q_s$, receive queue $Q_r$. \\
\hspace*{0.02in}  
\textbf{Output:} $u, w, s_{t}, b, v, c$.\\
\begin{algorithmic}[1]
\STATE $Q_r$ receives hyperparameters from the master.  
\STATE $Q_r$ receives $u, w, s_{t}, b, v, c$ from the master.
\STATE Assign all the parameters to $N_{ap}$ workers. 
\FOR{t = 1 to $K$} 
    \FOR{i = 1 to $N_{ap}$}
        \STATE worker $N_i$ updates its parameters.
        \STATE $Q_s$ sends those updated parameters to the master.
    \ENDFOR
	\STATE $Q_s$ sends $u, w, s_{t}, b, v, c$ to master.
\ENDFOR
\end{algorithmic}
\end{algorithm}

\subsection{Convergence Analysis}
In this section, we present the convergence analysis about ADMMiRNN. For convenience, we define $\rho=\{\rho_1, \rho_2, \rho_3\}$. First, we give some mild assumptions as follows:
\begin{figure*}[tb]
\centering
\subfigure[training loss versus iterations.]{
\includegraphics[width=0.46\textwidth]
{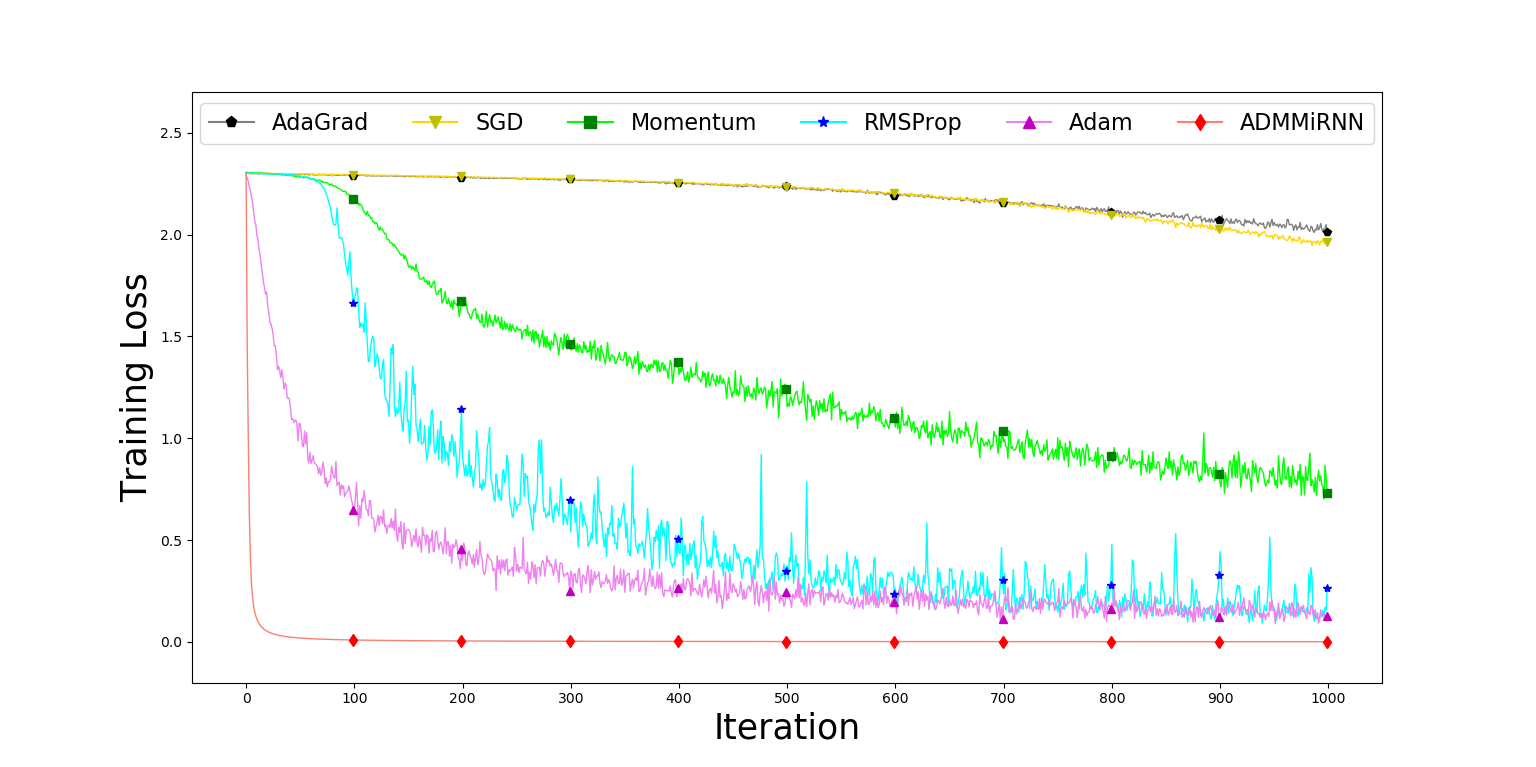}\label{fig:train_loss comparison}
}
\subfigure[test loss versus iterations.]{
\includegraphics[width=0.46\textwidth]
{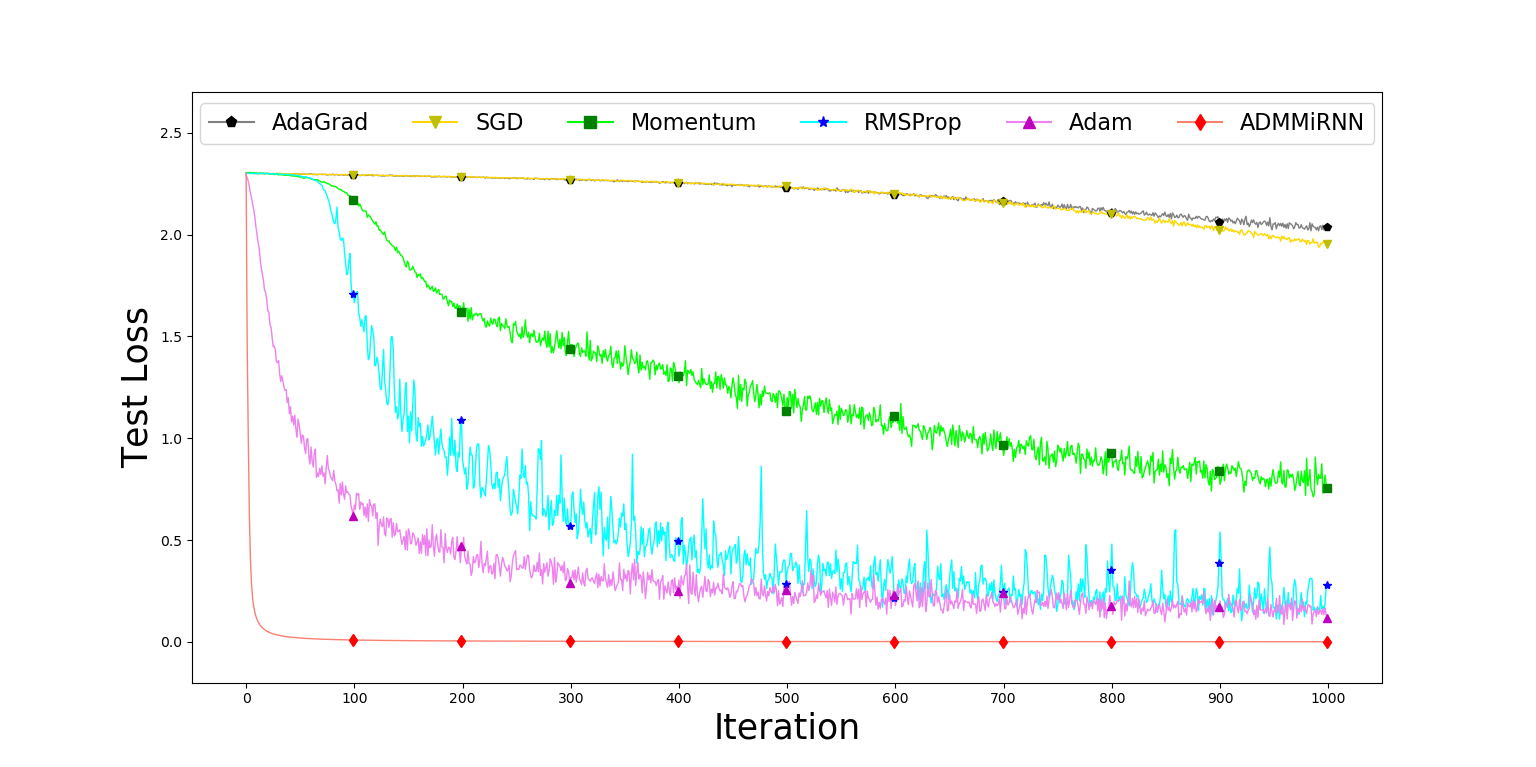}\label{fig:test_loss comparison}
}
\caption{Training loss and test loss versus iterations of ADMMiRNN, SGD, AdaGrad, Momentum, RMSprop, and Adam. ADMMiRNN achieves the best performance against other optimizers on MNIST.}\label{fig:loss comparison}
\end{figure*}
\begin{figure*}[tb]
\centering
\subfigure[training accuracy versus iterations.]{
\includegraphics[width=0.46\textwidth]
{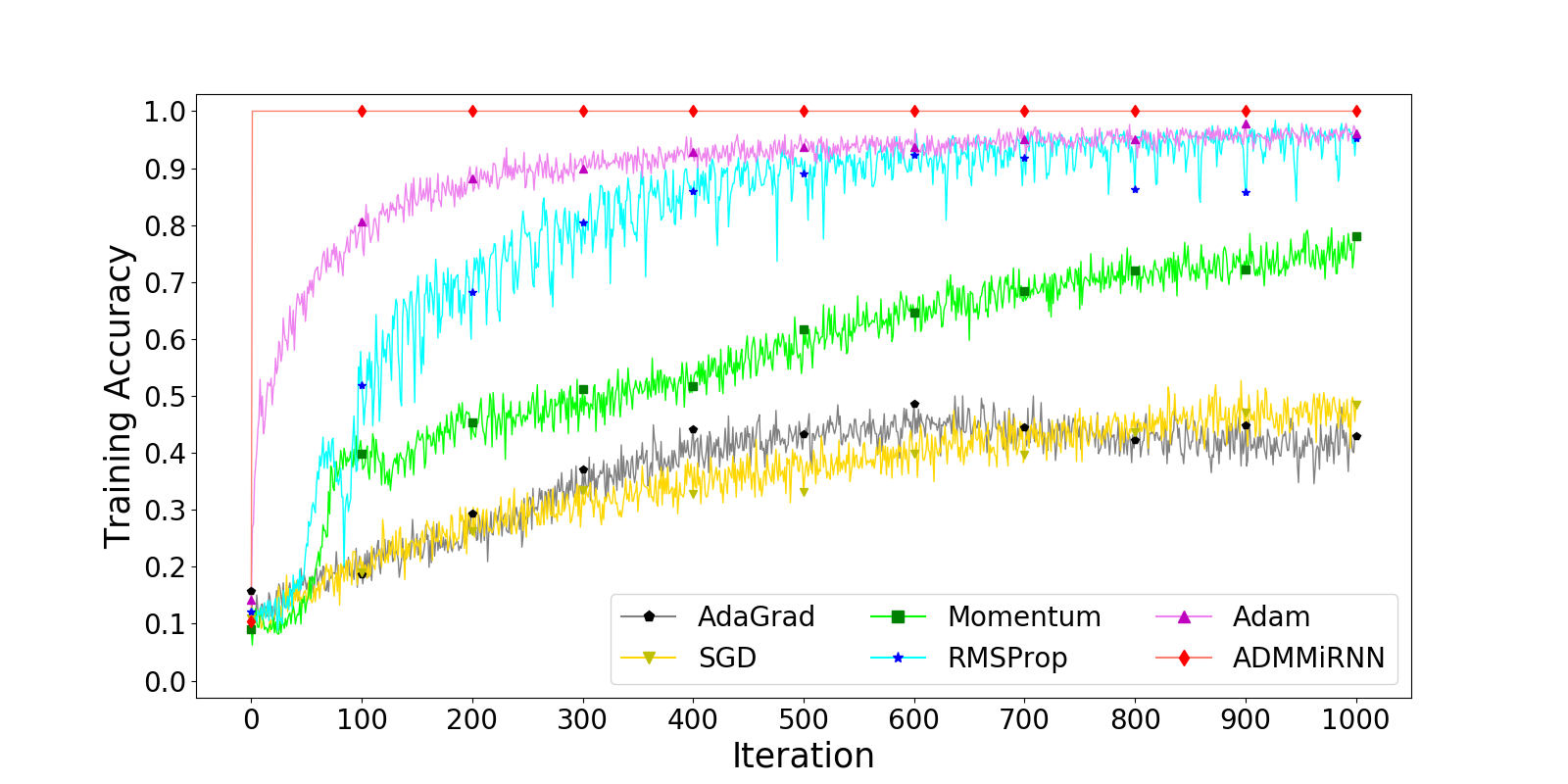}\label{fig:train accuracy}
}
\subfigure[test accuracy versus iterations.]{
\includegraphics[width=0.46\textwidth]
{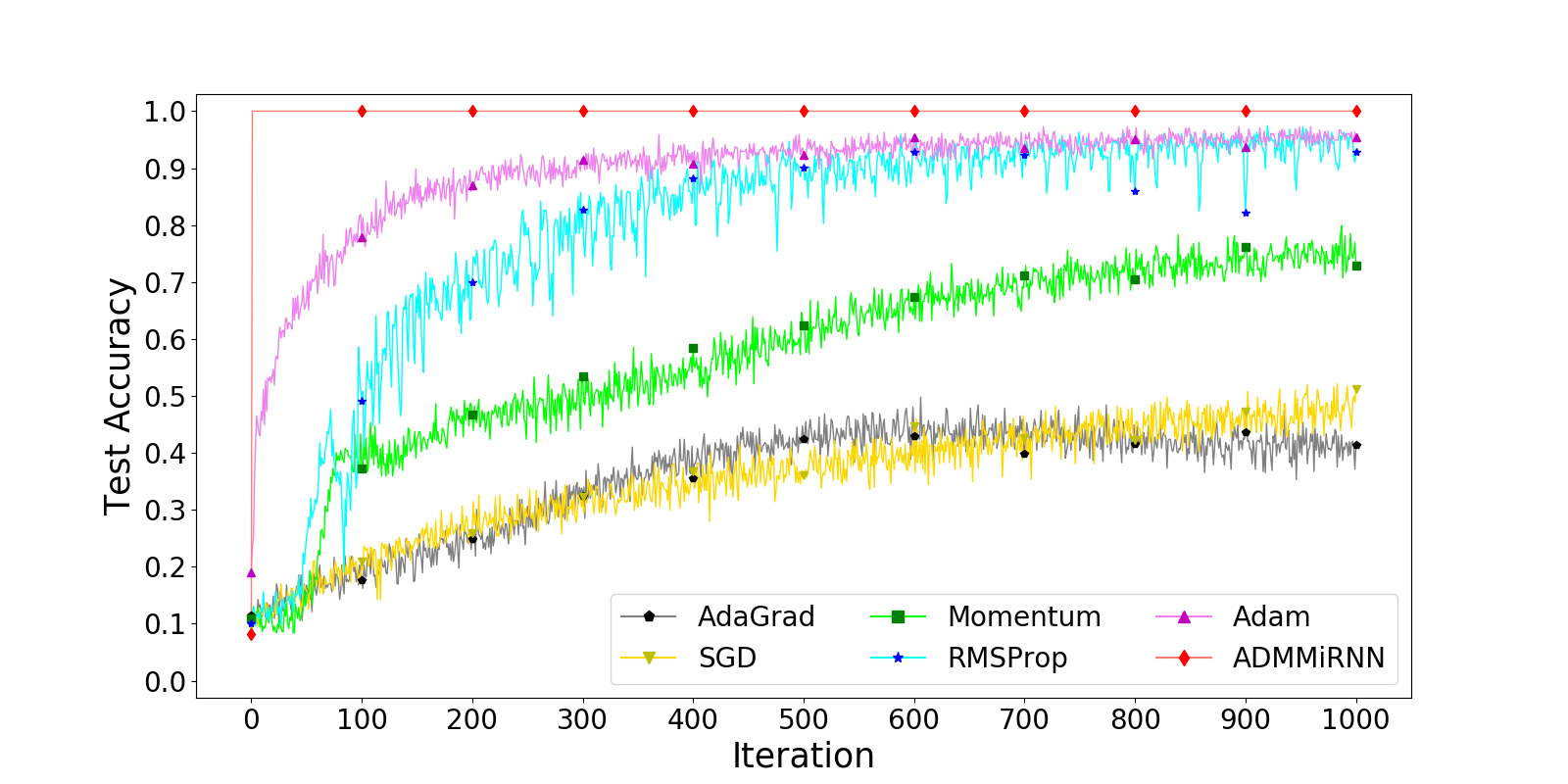}\label{fig:test accuracy}
}
\caption{Training accuracy and test accuracy versus iterations of ADMMiRNN, SGD, AdaGrad, Momentum, RMSprop, and Adam.}\label{fig:accuracy comparison}
\end{figure*}

{\bf Assumption 1.}\label{assumption:1} The gradient of $R$ is $H$-\textit{Lipschitz} continuous, $i.e.$, $\| \nabla R(o_1) - \nabla R(o_2)\| \le H\|o_1 - o_2\|$, $H \ge 0$ and is called the \textit{Lipschitz constant}. 
This is equivalent to $R(o_1) \le R(o_2) + \nabla R(o_2)\cdot (o_1 - o_2) + H/2\|o_1-o_2\|^2$;

{\bf Assumption 2.} The gradient of the objective function $\mathcal{L}_\rho$ is bounded, $i.e.$, there exists a constant $C$ such that $\nabla \mathcal{L}_\rho \le C$;

{\bf Assumption 3.} The second-order moment of the gradient $g_t$ is uniformly upper-bounded, that is to say $\mathbb{E}\|g_t\|^2 \le C$.

Such assumptions are typically used in~\cite{Saeed1,Saeed2,zou2019sufficient}.
Under these assumptions, we will have the properties~\cite{wang2019multi} shown in the supplementary materials. Then we can prove that ADMMiRNN converges under the following theorems.

{\bf Theorem 1.} If $\rho_i > 2H~(i=1,2,3)$ and \textbf{Assumption1-3} hold, then \textbf{Property 1-3} in the supplementary materials hold. 

{\bf Theorem 2.} If $\rho_i > 2H~(i=1,2,3)$, for the variables $(\theta, \lambda_1, \lambda_2, \lambda_3)$ in Problem~\ref{problem:2}, starting from any $(\theta^0, \lambda_1^0, \lambda_2^0, \lambda_3^0)$, it at least has a limit point $(\theta^*, \lambda_1^*, \lambda_2^*, \lambda_3^*)$ and any limit point $(\theta^*, \lambda_1^*, \lambda_2^*, \lambda_3^*)$ is a critical point of Problem~\ref{problem:2}. In other words, $0 \in \partial \mathcal{L}_{\rho_1,\rho_2,\rho_3}(\theta^*)$.

Theorem~2 concludes that ADMMiRNN has a global convergence. 

{\bf Theorem 3.} For a sequence $\theta$ generated by Algorithm~\ref{alg:RNN algorithm}, define $m_k=\min\limits_{0\le t \le k}(\|\theta^{\Tilde{k}}-\theta^k\|_2^2)$, the convergence rate of $m_k$ is $O(1/k)$.

Theorem~3 concludes that ADMMiRNN converges globally at a rate of $O(1/T)$. The convergence rate is consistent with the current work of ADMM~\cite{wang2019multi,zhong2014fast,ouyang2013stochastic}. 
Due to space limited, the proofs of the above theorems are also omitted in the supplementary materials.
This analysis is suitable for ADMMiRNN and SP-ADMMiRNN. 

When it comes to AP-ADMMiRNN, the convergence analysis is more complicated. 
This analysis is still based on former assumptions. From \cite{chang2016asynchronous}, we have the following lemma, 
\begin{figure*}[htbp]
\centering
\subfigure[Training loss of ADMMiRNN and some typical optimizers on IMDb.]{
\includegraphics[width=0.46\textwidth, height=14em]{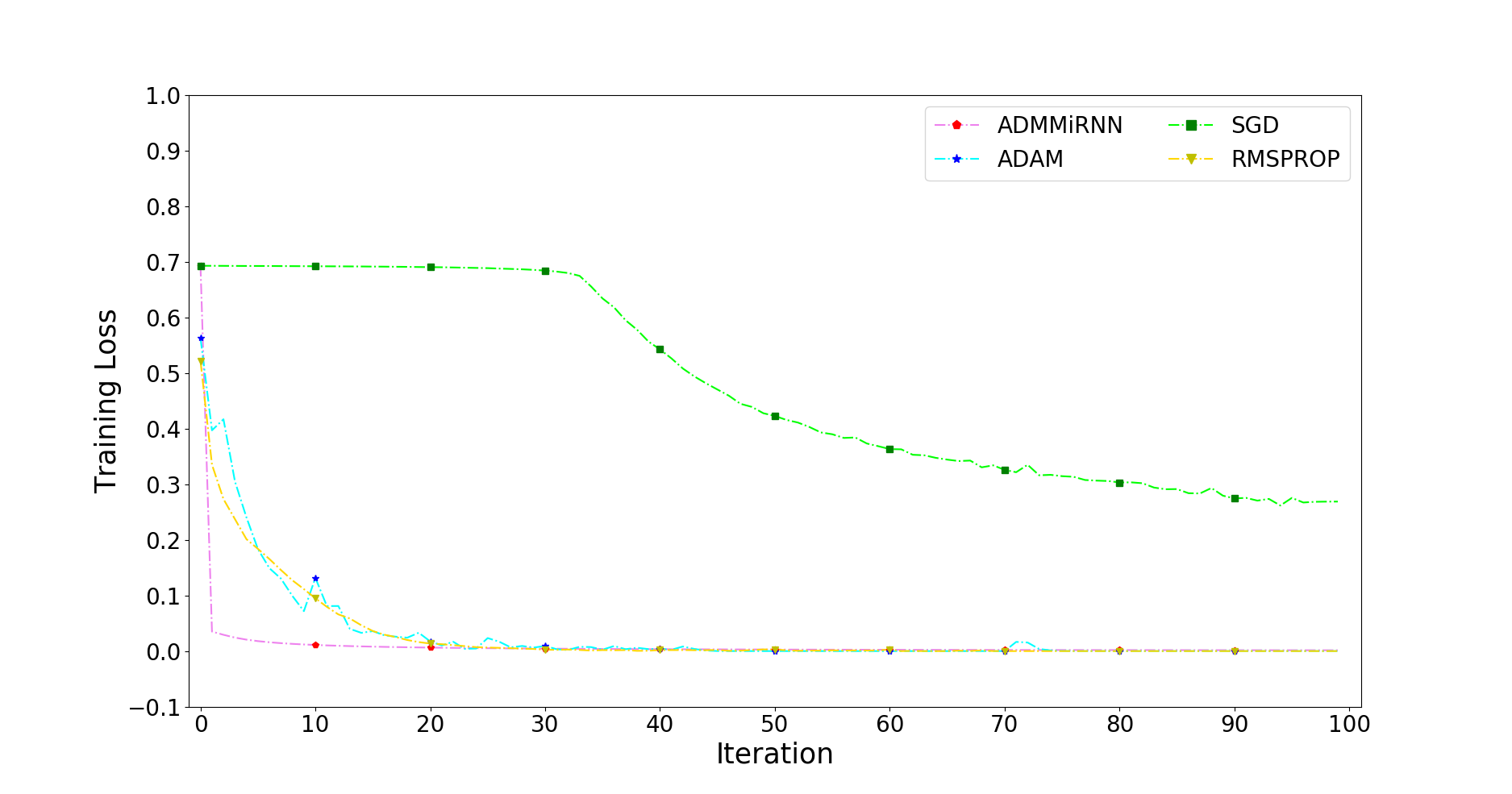}\label{fig:imdb loss}
}
\subfigure[Accuracy of ADMMiRNN and some typical optimizers on IMDb.]{
\includegraphics[width=0.46\textwidth, height=14em]{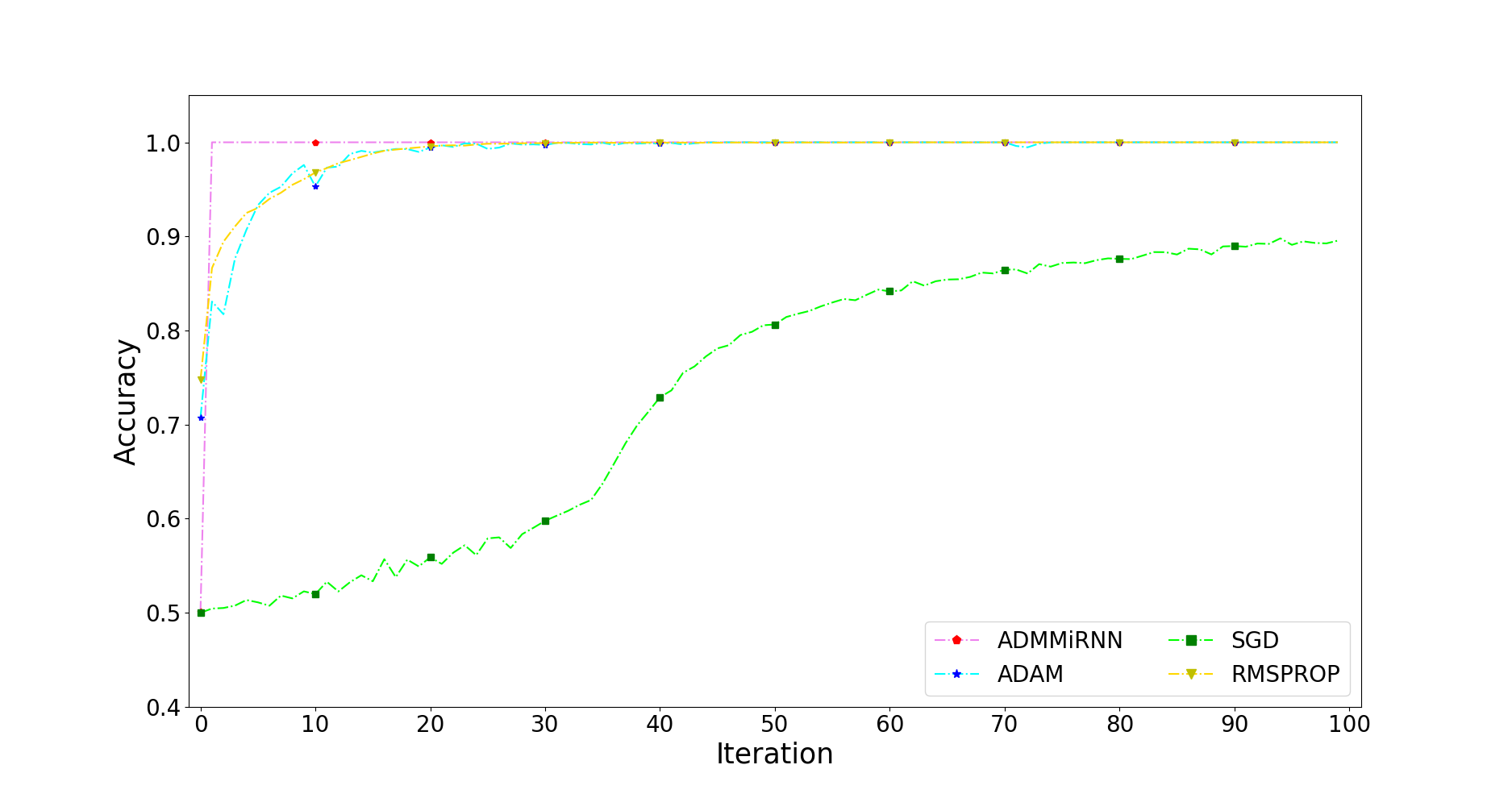}\label{fig:imdb acc}
}
\caption{The comparison of accuracy and loss among ADMMiRNN, SGD, Adam and RMSProp.}\label{fig:imdb}
\end{figure*}

{\bf Lemma 1.} There exists a constant $S \in [1, N]$ such that 
\begin{equation}
    \begin{aligned}
        \infty > & \mathcal{L}_{\rho}(\theta^{0}) - \Phi^{*} \ge 0,\\
        \rho_i > & \frac{(1+H+H^2)+\sqrt{(1+H+H^2)^2+8H^2}}{2} i=1,2,3,\\
    \end{aligned}
\end{equation}
where $\Phi^{*} > -\infty$ and is the optimal objective value of Problem~\ref{problem:1}. Then $\theta^t$ generated by Algorithm~\ref{alg:master} and Algorithm~\ref{alg:ap-admmirnn worker} are bounded and have limit points which satisfy KKT conditions of Problem~\ref{problem:2}.

Besides, in AP-ADMMiRNN, we need another additional assumption 4.

{\bf Assumption 4.} $R(o)$ is strongly convex with $\delta^2 > 0$.
\begin{figure*}[htbp]
\centering
\subfigure[Training loss of ADMMiRNN and some typical optimizers.]{
\includegraphics[width=0.46\textwidth, height=15em]{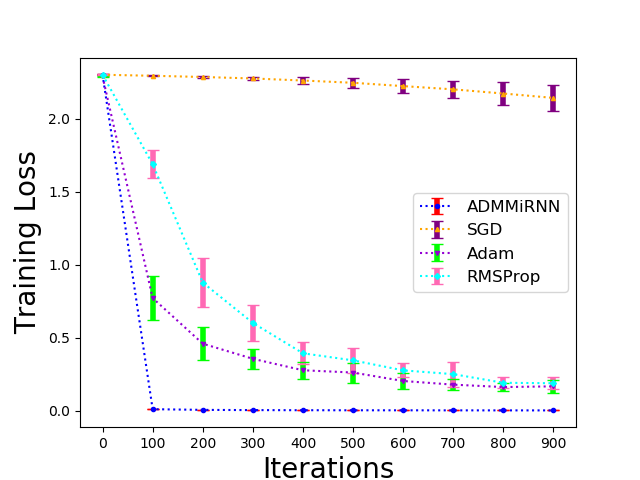}\label{fig:errbar_train}
}
\subfigure[test loss of ADMMiRNN and some typical optimizers.]{
\includegraphics[width=0.46\textwidth, height=15em]{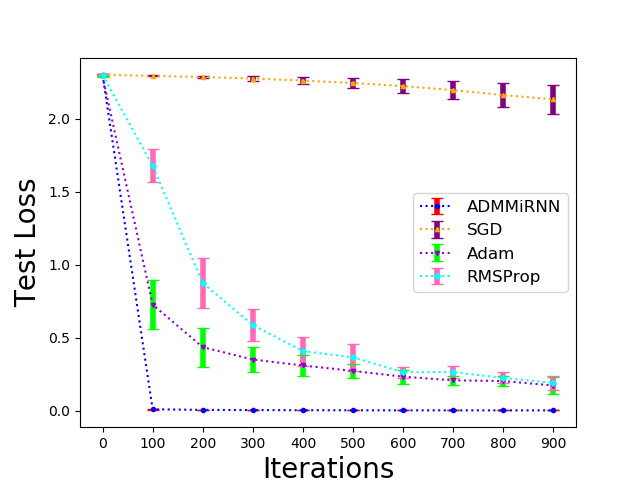}\label{fig:errbar_test}
}
\caption{The comparison of stability among ADMMiRNN, SGD, Adam and RMSProp. For each optimization method, we repeated experiments 10 times to obtain the mean and variance of the training loss and test loss against iterations on MNIST.}\label{fig:errorbar}
\end{figure*}

{\bf Theorem 4.} Let $\delta = 0$ and $0 < \rho_i \le \frac{\delta^2}{(5\tau -3)\max{2\tau, 3(\tau -1)}} (i=1,2,3)$ and $\theta^k$ is generated by Algorithm~\ref{alg:master} and Algorithm~\ref{alg:ap-admmirnn worker}. Then it holds that 
\begin{equation}
    \begin{aligned}
        \|\Phi(\theta^k) - \Phi(\theta)^{*}\| + \|\theta^k-\theta_0\| \le \frac{2+\sigma_{\lambda}C}{k}
    \end{aligned}
\end{equation}
Lemma 1 and Theorem 4 imply that Algorithm~\ref{alg:ap-admmirnn worker} is guaranteed to converge to the set of KKT points as long as $\rho_i (i=1,2,3)$ satisfy those conditions.
More details and proofs could be referred in \cite{chang2016asynchronous}.   

\section{Experiments}\label{sec:experiments}
\subsection{Setup}
We train a RNN model shown in Fig.\ref{fig:rnncell} on MNIST~\cite{lecun1998gradient} and IMDb~\cite{dodds2006popular}. 
This is achieved by NumPy, and those parameters are updated in a manner of Algorithm~\ref{alg:RNN algorithm}.
The MNIST dataset has 55,000 training samples and 10,000 test samples and was first introduced in~\cite{lecun1998gradient} to train handwritten-digit image recognition. The IMDb dataset consists of 50,000 movie reviews~(half negative and half positive). This dataset is split evenly into 25,000 reviews for training and 25,000 reviews for testing. 
All the experiments related to MNIST are conducted in 1000 iterations on a 64-bit Ubuntu 16.04 system.

Furthermore, our experiments are also conducted on a text.
The text could also be accessed from our open-source code repository.
Training on a text is a typical RNN task.
We achieved a typical RNN model and unfolded it to $N$ cells with NumPy and $N$ is also the length of the input sequence. 
In our experiments, we adopt a kind of smooth loss.
These experiments are performed on a Macbook Pro with an Intel 3.1~GHz Core i5 Processor and 8~GB Memory. 

In our paralleled experiments, we train ADMMiRNN, SP-ADMMiRNN, and AP-ADMMiRNN on MNIST in 30 iterations. Those experiments are performed on an Ubuntu-16 system with two 1080-Ti GPUs. 

We utilize a fixed value strategy for these hyperparameters in all of our experiments, such as $\rho_1, \rho_2$, and $\rho_3$. 

\subsection{Convergence Results}
\subsubsection{Results on MNIST}\label{subsubsec:results on mnist}
We train the simple RNN model shown in Fig.\ref{fig:rnncell} through different optimizers, including SGD, Adam, Momentum~\cite{qian1999momentum}, RMSProp~\cite{tieleman2012lecture} and AdaGrad~\cite{duchi2011adaptive}.
We compare our ADMMiRNN with these commonly-used optimizers in the loss and accuracy and display our experimental results on MNIST in Fig.\ref{fig:loss comparison} and Fig.~\ref{fig:accuracy comparison} respectively.
Both Fig.~\ref{fig:loss comparison} and Fig.\ref{fig:accuracy comparison} indicate that ADMMiRNN converges faster than the other optimziers.
ADMMiRNN gets a smoother loss curve while the loss curves of other optimizers shake a lot.
This means ADMMiRNN trains models in a relatively stable process.
Besides, ADMMiRNN gets much lower training loss and test loss as well as promising training and test accuracy.
These results prove that ADMMiRNN could converge in RNN tasks and confirm that ADMMiRNN is a much more powerful tool than traditional gradient-based optimizers in deep learning. 

\subsubsection{Results on IMDb}
Fig.\ref{fig:imdb} shows that our additional experiments comparing the training loss and accuracy on IMDb of ADMMiRNN, SGD, Adam, and RMSProp. In this figure, we find that ADDMiRNN converges faster than SGD and Adam and RMSProp, which is consistent with the trends in Fig.~\ref{fig:loss comparison} and Fig.~\ref{fig:accuracy comparison}. ADMMiRNN achieves a similar result as Adam and RMSProp do considering the training loss. As for the accuracy, ADMMiRNN reaches the top value faster than the others. This is also consistent with the results in Section~\ref{subsubsec:results on mnist}. In total, ADMMiRNN behaves better than these typical optimizers.

In our experiments on MNIST and IMDb, we find the accuracy of ADMMiRNN always reaches 1.0 within several iterations. According to our analysis, we choose to solve the target directly instead of computing gradients in ADMM, which speeds up the convergence and produces an intuitively better solution. 

\begin{figure}
    \centering
    \includegraphics[width=0.46\textwidth, height=14em]{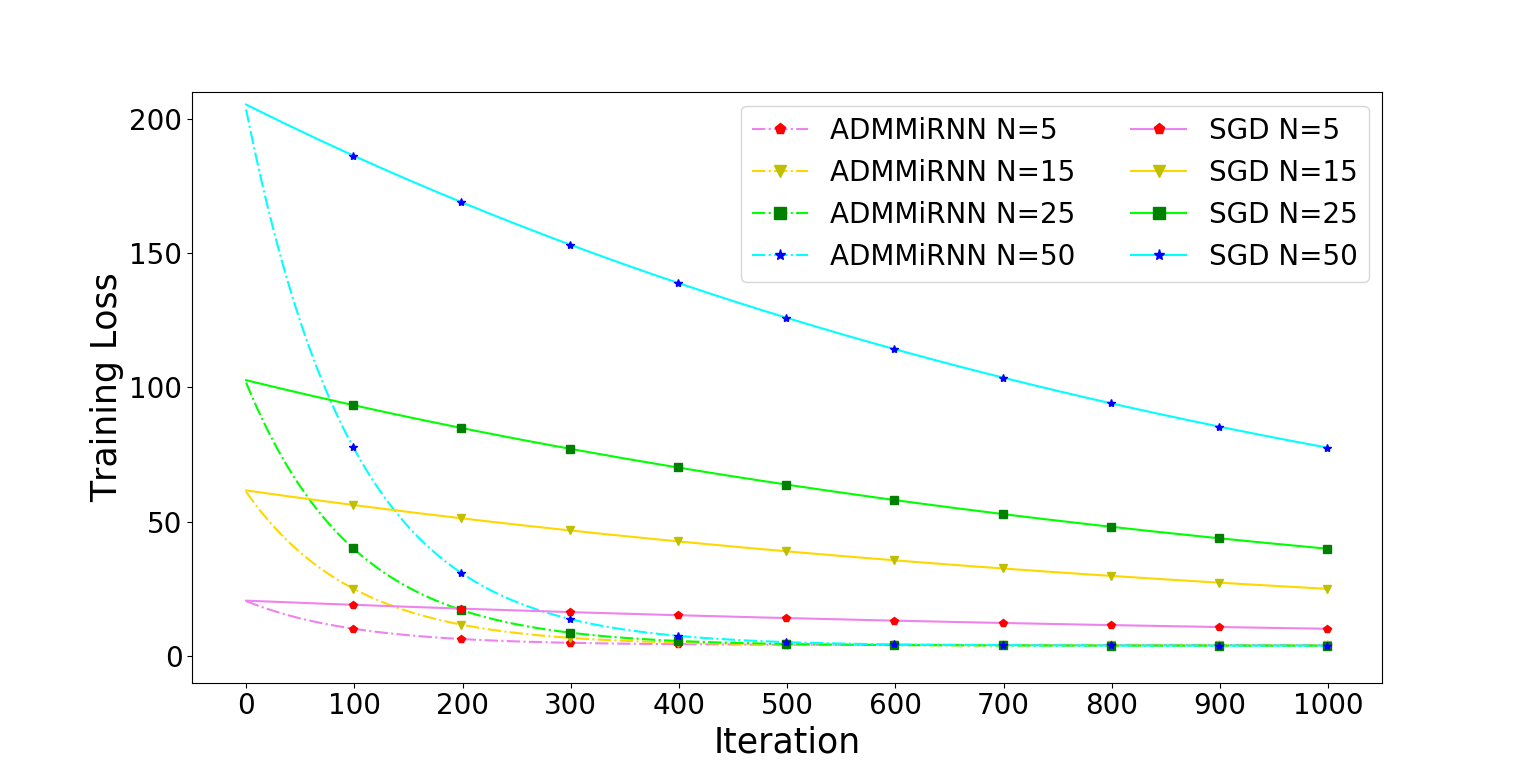}
    \caption{The results of ADMMiRNN and SGD on different input sequence length. In this figure, $N$ represents the length.}
    \label{fig:length}
\end{figure}
\subsubsection{Results on Text Data}
Besides experiments on MNIST, we also explore how ADMMiRNN performs in text classification tasks.
One critical shortcoming of current RNN models is that they are sensitive to the length of the input sequence because the longer the input sequence is, the worse training results are. To investigate the sensitivity of ADMMiRNN to the input lengths, we measure the performance of ADMMiRNN and SGD on the text data with different input sequence length. The results are displayed in Fig.~\ref{fig:length}. Here, we adopt the average loss of the input sequence as our target. From Fig.~\ref{fig:length}, we have evidence that ADMMiRNN always produces a remarkable result and is nearly immune to the length, which performs much more impressive than SGD regardless of the length of the input sequence.  


\begin{table}[tb]
\centering
\caption{Training loss and test loss under different hyperparameter settings. All of these values are obtained after 20 iterations.}
\begin{tabular}{cccccc}  
\toprule
$\rho_1$ & $\rho_2$ & $\rho_3$ & $r$ & training loss & test loss  \\
\midrule
1 & 1 & 1 & 1       & $5.045\times10^{-2}$  & $5.046\times10^{-2}$   \\
0.1 & 1 & 1 & 1    & $5.339\times10^{-2}$  & $5.338\times10^{-2}$   \\
1 & 0.1 & 1 & 1     & $5.338\times10^{-2}$      & $5.340\times10^{-2}$     \\
1 &1 & 0.1 & 1   & $3.776\times10^{-4}$  & $3.776\times10^{-4}$      \\
1 &1 & 10  & 1            & $0.9984$  & $0.9985$     \\
1 & 1 & 1 & 10   &  $5.339\times10^{-2}$     &  $5.338\times10^{-2}$    \\
1 & 1 & 10 & 10         & $0.9987$   & $0.9986$      \\
\bottomrule
\end{tabular}
\label{tab:hyper loss}
\end{table}

\subsection{Stability}
As aforementioned, initial weights and biases are critical in RNN models. 
In this section, we mainly compare ADMM with some different optimizers and explore its stability for RNN.
In brief, we compare ADMMiRNN with SGD, Adam, and RMSProp and repeat each scheme ten times independently.
The experimental results are displayed in Fig.\ref{fig:errorbar}.
The blocks in Fig.~\ref{fig:errbar_train} and Fig.~\ref{fig:errbar_test} represent the standard deviation of the samples drawn from the training and testing process.
The smaller the blocks are, the more stable the method is. 
From Fig.\ref{fig:errbar_train} and Fig.\ref{fig:errbar_test}, we observe that at the beginning, SGD has a small fluctuation. Nevertheless, as the training progresses, the fluctuation gets more and sharper, which means that SGD tends to be unstable. 
As for Adam and RMSProp, their variance is smaller but still significant about ADMMiRNN. 
According to different initialization of weights and biases, these optimizers may cause different results within a big gap between them.
Specifically, ADMMiRNN has a relatively small variance from beginning to end compared with SGD, Adam, and RMSProp, which is too small to show clearly in Fig.\ref{fig:errbar_train} and Fig.\ref{fig:errbar_test}, which indicates that ADMMiRNN is immune to the initialization of weights and biases and settle the sensitivity of RNN models to initialization.

\begin{figure}
    \centering
    \includegraphics[width=0.46\textwidth, height=14em]{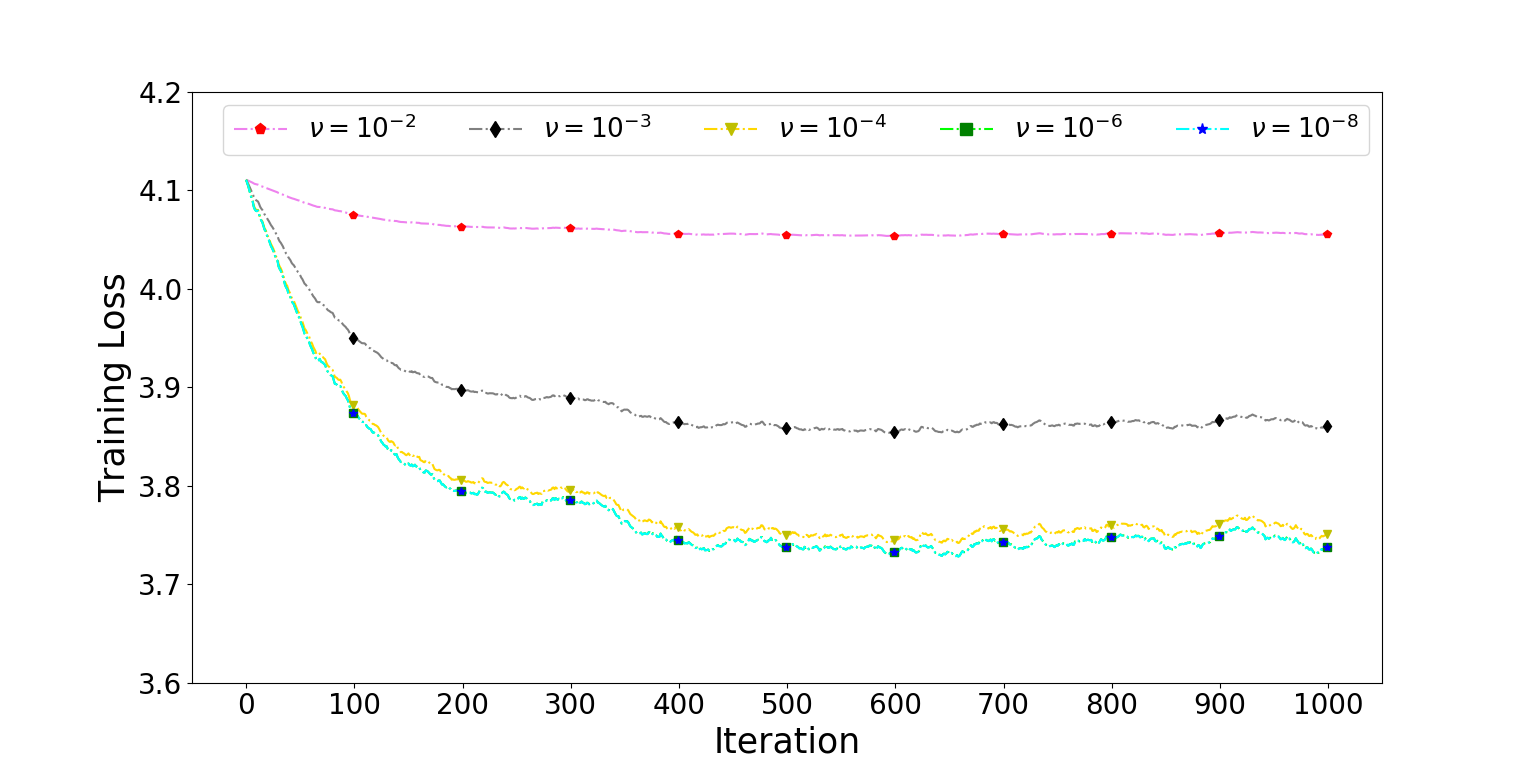}
    \caption{The training loss V.S. iterations of ADMMiRNN on a text classification task with different $\nu$.}
    \label{fig:nu}
\end{figure}

No matter how the initialization changes, ADMMiRNN always gives a stable training process and promising results.
The results demonstrate that ADMMiRNN is a more stable training algorithm for RNN models than stochastic gradient algorithms.
\begin{figure}
    \centering
    \includegraphics[width=0.46\textwidth, height=14em]{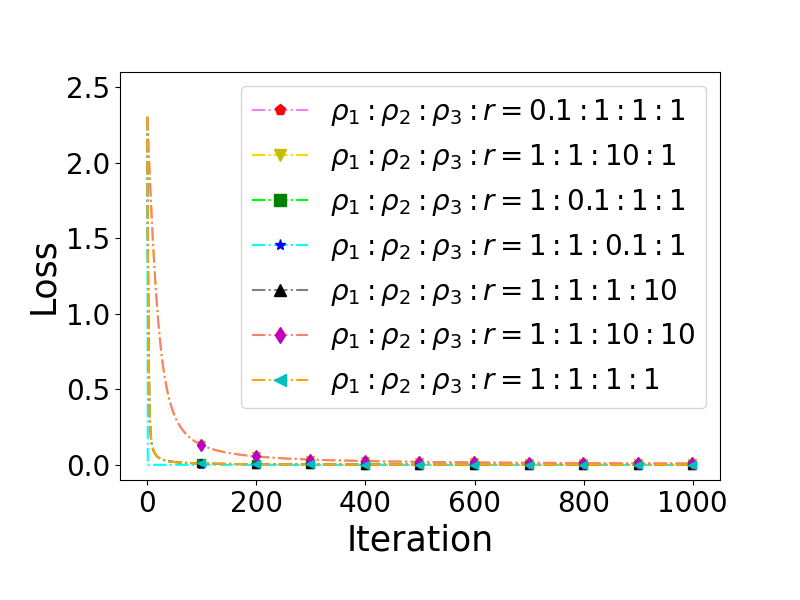}
    \caption{The losses values V.S. iteration with different values of $\rho$s.}
    \label{fig:hyper}
\end{figure}

\subsection{Choices of Hyperparameters $\rho$s and $\nu$}
\subsubsection{varying $\rho$s}
In vanilla ADMM, the value of the penalty term is critical, and it may have adverse effects on convergence.
In this subsection, we mainly try different hyperparameters in ADMMiRNN and evaluate how they influence the training process of ADMMiRNN. 
These results are summarized in Table~\ref{tab:hyper loss} and Fig.~\ref{fig:hyper}.
Table~\ref{tab:hyper loss} implies that $\rho_3$ determines the best result in ADMMiRNN.
More precisely, we find that larger $\rho_3$ delays the convergence speed in ADMMiRNN from Fig.~\ref{fig:hyper}.
However, if $\rho_3$ is too large, it may produce non-convergent results.
Furthermore, it turns out that $\rho_1$ and $\rho_2$ account less in ADMMiRNN while $\rho_3$ plays a much more crucial role with regard to the property of convergence and its convergence speed.

\subsubsection{varying $\nu$}
In this subsection, we investigate the influence of $\nu$ in Eq.~\eqref{equ:phi}. In our experiments on a text data, we fix all the hyperparameters other than $\nu$ and set it $10^{-2}$, $10^{-3}$, $10^{-4}$, $10^{-6}$, $10^{-8}$ respectively. We display the curves corresponding to different values of $\nu$ in Fig.~\ref{fig:nu}.
Fig.~\ref{fig:nu} suggests that larger $\nu$ produces a relatively worse convergence result in ADMMiRNN. Small $\nu$ can not only lead to a slight loss but is also able to push the training process to converge fast. However, when $\nu$ is small enough, the influence on the convergence rate and convergent result is not apparent.   

\subsection{Paralleled Experiments}
\begin{figure}
    \centering
    \includegraphics[width=0.46\textwidth, height=14em]{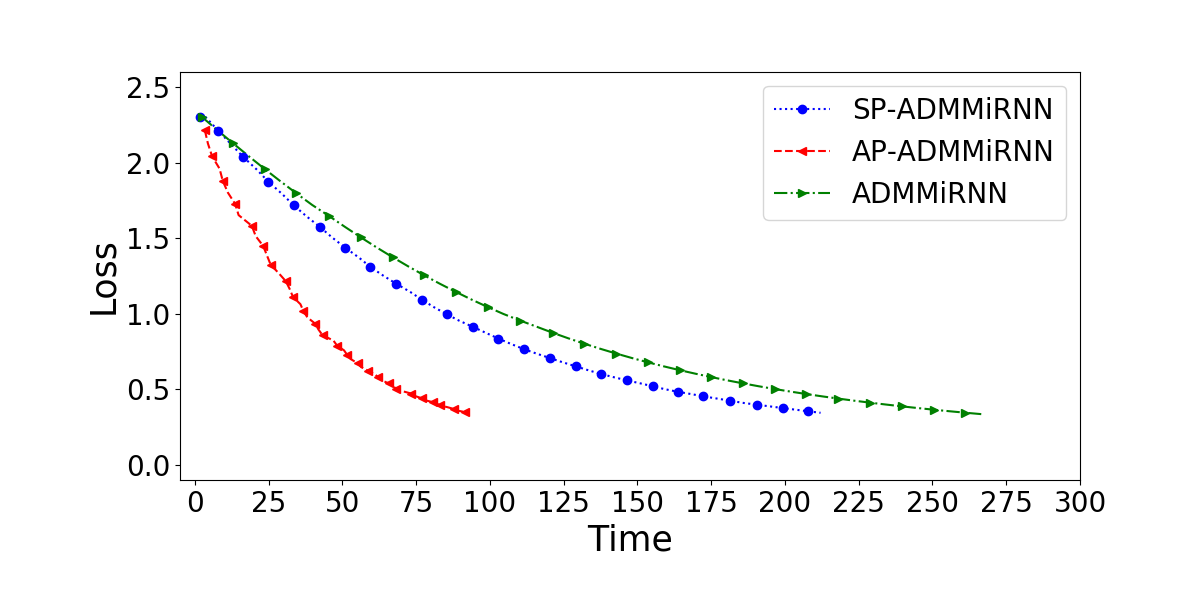}
    \caption{The loss values V.S. time of vanilla ADMMiRNN, SP-ADMMiRNN and AP-ADMMiRNN.}
    \label{fig:comparasion of 3 workers}
\end{figure}

\begin{figure}
    \centering
    \includegraphics[width=0.4\textwidth, height=12em]{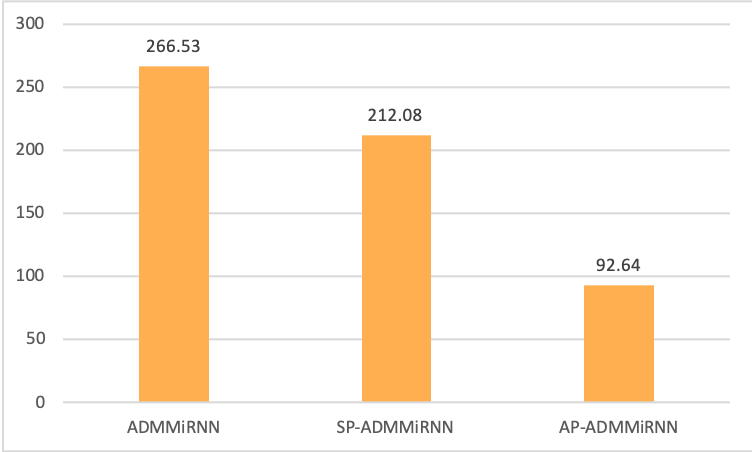}
    \caption{Comparison of training time of vanilla ADMMiRNN, SP-ADMMiRNN and AP-ADMMiRNN within 50 iterations. The AP-ADMMiRNN is much faster than the others.}
    \label{fig:cost time 3 workers}
\end{figure}

To compare vanilla ADMMiRNN, SP-ADMMiRNN and AP-ADMMiRNN, we conduct several experiments on MNIST and present the experimental results in Fig.~\ref{fig:comparasion of 3 workers} and Fig.~\ref{fig:cost time 3 workers}. In this test, there are 3 workers in both SP-ADMMiRNN and AP-ADMMiRNN. SP-ADMMiRNN and AP-ADMMiRNN take less time than vanilla ADMMiRNN.
The vanilla ADMMiRNN costs 266.53 seconds in 50 iterations, nearly one minute more than the time of SP-ADMMiRNN. Therefore, SP-ADMMiRNN could solve not only large-scale models that cannot be solved on a single machine through distributed algorithms but also can save time.  

Comparing AP-ADMMiRNN with the vanilla ADMMiRNN and SP-ADMMiRNN, it takes the least time in these three methods, which is about one-third of that of the vanilla ADMMiRNN. This is a great speedup given 3 workers in the system. It is also worth noting that AP-ADMMiRNN converges faster than both ADMMiRNN and SP-ADMMiRNN. 

\section{Conclusion}\label{sec:conclusion}
In this paper, we proposed a new framework to train RNN tasks, namely ADMMiRNN. 
Since it is challenging to train RNNs with ADMM directly, we set up ADMMiRNN on the foundation of the expanded form of RNNs. 
The convergence analysis of ADMMiRNN is presented, and ADMMiRNN could achieve a convergence rate of $O(1/T)$. 
We further conduct several experiments on real-world datasets based on our theoretical analysis. 
Experimental results of comparisons regarding ADMMiRNN and several popular optimizers manifest that ADMMiRNN converges faster than these gradient-based optimizers.
Besides, it presents a much more stable process than them.
To the best of our knowledge, we are the first to apply ADMM into RNN tasks and present theoretical analysis, and ADMMiRNN is the first to alleviate the \textit{vanishing} and \textit{exploding} gradients problem and the sensitivity of RNN models to initializations at the same time.
In conclusion, ADMMiRNN is a promising tool to train RNN models. 
Another important contribution of our work is P-ADMMiRNN, including Synchronous Parallel ADMMiRNN and Asynchronous Parallel ADMMiRNN. 
This is a new form of model parallelism, and P-ADMMiRNN is much easier to be realized.  
We train ADMMiRNN in both synchronous and asynchronous parallel ways and make a fair comparison among vanilla ADMMiRNN, SP-ADMMiRNN, and AP-ADMMiRNN. 
Experiments demonstrate that AP-ADMMiRNN converges faster than vanilla ADMMiRNN and SP-ADMMiRNN.
Further experiments illustrate that the number of workers is also critical concerning the speedup ratio, which means we cannot make the master too busy for better training performance.
In the future, we will explore how ADMMiRNN performs in large datasets.

%

\appendices


\section{Update $w$}
As for the update of $w$ in Eq.~\eqref{equ:lagrangian}, 
at iteration $k$, it is updated as follows:
\begin{equation}\nonumber
    w^{\Tilde{k}} \leftarrow \arg\min \mathcal{L}_{\rho_1,\rho_2,\rho_3}(u^{\Tilde{k}},w,b^k,a^k,s^k,v^k,c^k,o^k).
\end{equation}
which is equivalent to the following form:
\begin{equation}\label{equ:update_wt}
    w^{\Tilde{k}} \leftarrow \arg\min \Omega(w) + \phi(u^{\Tilde{k}},w,b^k,a^k,s^k,v^k,c^k,o^k).
\end{equation}

\section{Update $u$}
Similar as the update of $u$ in Section~\ref{sec:update_u}, 
we also define $\textbf{G}=rI_{d}-\rho_1 s_{t-1}^Ts_{t-1}$ and use with linearized proximal point method, then the update of $w$ is transformed into
\begin{equation}
    \begin{aligned}\label{equ:update_w}
        w^{\Tilde{k}} \leftarrow & \arg\min\limits_{w} \Omega(w) \frac{Nr}{2}\|w-w^k\|^2+\nu(w-w^k)^T\sum_{t=1}^{N-1}[(s_{t-1}^k)^T\\
        & (a_t^k-u^kx_t^k-w^ks_{t-1}^k-b^k)]+\rho_1(w-w^k)^T[(s_{N-1}^k)^T  \\
        & (a_N^k-u^kx_N^k-w^ks_{N-1}^k-b^k-\lambda_1^k/\rho_1)].  \\
    \end{aligned}
\end{equation}

\section{Update $b$}
As far as $b$ is concerned, it 
has a similar updating rule.
\begin{equation}\nonumber
    b^{\Tilde{k}} \leftarrow \arg\min \mathcal{L}_{\rho_1,\rho_2,\rho_3}(u^{\Tilde{k}},w^{\Tilde{k}},b,a^k,s^k,v^k,c^k,o^k),
\end{equation}
and the updating rule of $b$ is transformed into:
\begin{equation}\label{equ:update_bt}
    b^{\Tilde{k}} \leftarrow \arg\min\limits_{b} \phi(u^{\Tilde{k}},w^{\Tilde{k}},b,a^k,s^k,v^k,c^k,o^k).
\end{equation}

\section{Update $v$}
We spot that $v$ and $s_t$ are not decoupled in Eq.~\eqref{equ:phi}.
Therefore, 
to avoid high computational complexity, we adopt a similar way as that in updating $u$ and $w$. 
The parameter $v$ is updated as follows:
\begin{equation}\nonumber
    v^{\Tilde{k}} \leftarrow \arg\min \mathcal{L}_{\rho_1,\rho_2,\rho_3}(u^{\Tilde{k}},w^{\Tilde{k}},b^{\Tilde{k}},a^{\Tilde{k}},s^{\Tilde{k}},v,c^k,o^k),
\end{equation}
Equally, adapt the following form and update $v_t$.
\begin{equation}\nonumber
    v^{\Tilde{k}} \leftarrow \arg\min \phi(u^{\Tilde{k}},w^{\Tilde{k}},b^{\Tilde{k}},a^{\Tilde{k}},s^{\Tilde{k}},v,c^k,o^k).
\end{equation}
Similar as aforementioned, the update rule for $v$ is 
\begin{equation}\label{equ:update_vt}
    \begin{aligned}
        v^{\Tilde{k}} \leftarrow & \arg\min\limits_{v} \frac{Nr}{2}\|v-v^k\|^2+\nu(v-v^k)^T\sum_{t=1}^{N-1}[(s_t^k)^T(o_t^k\\
        &  - v^ks_t^k-c^k)]+\rho_3(v-v^k)^T[(s_N^k)^T(o_N^k-v^ks_N^k-    \\
        & c^k -\lambda_1^k/\rho_3)].  \\
    \end{aligned}
\end{equation}

\section{Update $c$}
The parameter $c$ is quite simple, which is updated as follows:
\begin{equation}\nonumber
    c^{\Tilde{k}} \leftarrow \arg\min \mathcal{L}_{\rho_1,\rho_2,\rho_3}(u^{\Tilde{k}},w^{\Tilde{k}},b^{\Tilde{k}},a^{\Tilde{k}},s^{\Tilde{k}},v^{\Tilde{k}},c,o^k),
\end{equation}
which is equivalent to the following form:
\begin{equation}\label{equ:update_ct}
    c^{\Tilde{k}} \leftarrow \arg\min\limits_{c} \phi(u^{\Tilde{k}},w^{\Tilde{k}},b^{\Tilde{k}},a^{\Tilde{k}},s^{\Tilde{k}},v^{\Tilde{k}},c,o^k).
\end{equation}

\section{Update $o$}
Finally, we update $o_t$.
through:
\begin{equation}\nonumber
    o_t^{\Tilde{k}} \leftarrow \arg\min R(o) + \phi(u^{\Tilde{k}},w^{\Tilde{k}},b^{\Tilde{k}},a^{\Tilde{k}},s^{\Tilde{k}},v^{\Tilde{k}},c^{\Tilde{k}},o_t).
\end{equation}
It has to be noted that each $o_t$ is also updated separably. If $t<N$, 
\begin{equation}\label{equ:update_ot}
    \begin{aligned}
        o_t^{\Tilde{k}} \leftarrow \arg\min\limits_{o_t} R(o)+\frac{\nu}{2}\|o_t^k-v^ks_t^k-c^k\|^2 .
    \end{aligned}
\end{equation}
If $t=N$,
\begin{equation}\label{equ:update_oN}
    \begin{aligned}
        o_N^{\Tilde{k}} \leftarrow \arg\min\limits_{o_N} R(o)+\frac{\rho_3}{2}\|o_N^k-v^ks_N^k-c^k-\lambda_3^k/\rho_3\|^2 .
    \end{aligned}
\end{equation}

\ifCLASSOPTIONcaptionsoff
  \newpage
\fi


\begin{thebibliography}{1}

\bibitem{tang2020admmirnn}
Y Tang. et al. (2021) ADMMiRNN: Training RNN with Stable Convergence via an Efficient ADMM Approach. In: Hutter F., Kersting K., Lijffijt J., Valera I. (eds) Machine Learning and Knowledge Discovery in Databases. ECML PKDD 2020. Lecture Notes in Computer Science, vol 12458. Springer, Cham. https://doi.org/10.1007/978-3-030-67661-2\_1

\bibitem{bengio1994learning}
Bengio, Y., Simard, P., Frasconi, P., et al.: Learning long-term dependencies with gradient descent is difficult. IEEE transactions on neural networks \textbf{5}(2), 157--166 (1994)

\bibitem{boyd2011distributed}
Boyd, S., Parikh, N., Chu, E., Peleato, B., Eckstein, J., et al.: Distributed optimization and statistical learning via the alternating direction method of multipliers. Foundations and Trends{\textregistered} in Machine learning \textbf{3}(1), 1--122 (2011)

\bibitem{duchi2011adaptive}
Duchi, J., Hazan, E., Singer, Y.: Adaptive subgradient methods for online learning and stochastic optimization. Journal of Machine Learning Research \textbf{12}(Jul), 2121--2159 (2011)

\bibitem{elman1990finding}
Elman, J.L.: Finding structure in time. Cognitive science \textbf{14}(2), 179--211 (1990)

\bibitem{gabay1983augmented}
Gabay, D.: Augmented lagrangian methods: applications to the solution of boundary-value problems, chapter applications of the method of multipliers to
variational inequalities. North-Holland, Amsterdam \textbf{3}, 4 (1983)

\bibitem{mikolov2010recurrent}
Kombrink, S., Mikolov, T., Karafi${\acute a}$t, M., \& Burget, L. (2011). 
Recurrent neural network based language modeling in meeting recognition. In {\it Twelfth annual conference of the international speech communication association}.

\bibitem{gabay1976dual}
Gabay, D., Mercier, B.: A dual algorithm for the solution of nonlinear variational problems via finite element approximation. Computers \& mathematics with applications \textbf{2}(1), 17--40 (1976)


\bibitem{glowinski1989augmented}
Glowinski, R., Le Tallec, P.: Augmented Lagrangian and operator-splitting methods in nonlinear mechanics, vol. 9. SIAM (1989)

\bibitem{goldfarb2014robust}
Goldfarb, D., Qin, Z.: Robust low-rank tensor recovery: Models and algorithms. SIAM Journal on Matrix Analysis and Applications \textbf{35}(1), 225--253 (2014)

\bibitem{goodfellow2016deep}
Goodfellow, I., Bengio, Y., Courville, A.: Deep learning. MIT press (2016)

\bibitem{graves2007multi}
Graves, A., Fern{\'a}ndezz, S., Schmidhuber, J.: Multi-dimensional recurrent neural networks. In: International conference on artificial neural networks. pp. 549--558. Springer (2007)

\bibitem{hochreiter1997long}
Hochreiter, S., Schmidhuber, J.: Long short-term memory. Neural computation \textbf{9}(8), 1735--1780 (1997)

\bibitem{Kingma2014Adam}
Kingma, D., Ba, J.: Adam: A method for stochastic optimization. Computer Science (2014)

\bibitem{krizhevsky2012imagenet}
Krizhevsky, A., Sutskever, I., Hinton, G.E.: Imagenet classification with deep convolutional neural networks. In: Advances in neural information processing systems. pp. 1097--1105 (2012)

\bibitem{Lai2015TC}
Lai, S., Xu, L., Liu, K., Zhao, J.: Recurrent convolutional neural networks for text classification. In: AAAI. vol. 333, pp. 2267--2273 (2015)

\bibitem{lecun2015deep}
LeCun, Y., Bengio, Y., Hinton, G.: Deep learning. nature \textbf{521}(7553), 436--444 (2015)

\bibitem{lecun1998gradient}
LeCun, Y., Bottou, L., Bengio, Y., Haffner, P., et al.: Gradient-based learning applied to document recognition. Proceedings of the IEEE \textbf{86}(11), 2278--2324 (1998)

\bibitem{monteiro2010iteration}
Monteiro, R.D., Svaiter, B.F.: Iteration-complexity of block-decomposition algorithms and the alternating minimization augmented lagrangian method. Manuscript, School of Industrial and Systems Engineering, Georgia Institute of Technology, Atlanta, GA pp. 30332--0205 (2010)

\bibitem{nair2010rectified}
Nair, V., Hinton, G.E.: Rectified linear units improve restricted boltzmann machines. In: Proceedings of the 27th international conference on machine learning
(ICML-10). pp. 807--814 (2010)

\bibitem{Nguyen2016JRNN}
Nguyen, T.H., Cho, K., Grishman, R.: Joint event extraction via recurrent neural
networks. In: Proceedings of the 2016 Conference of the North American Chapter of the Association for Computational Linguistics: Human Language Technologies. pp. 300--309 (2016)

\bibitem{pascanu2013difficulty}
Pascanu, R., Mikolov, T., Bengio, Y.: On the difficulty of training recurrent neural networks. In: International conference on machine learning. pp. 1310--1318 (2013)

\bibitem{qian1999momentum}
Qian, N.: On the momentum term in gradient descent learning algorithms. Neural
networks \textbf{12}(1), 145--151 (1999)

\bibitem{huo2018decoupled}
Huo, Z., Gu, B., Yang, Q., \& Huang, H. (2018). Decoupled parallel backpropagation with convergence guarantee. arXiv preprint arXiv:1804.10574.

\bibitem{robbins1951stochastic}
Robbins, H., Monro, S.: A stochastic approximation method. The annals of math-
ematical statistics pp. 400--407 (1951)

\bibitem{rockafellar1976monotone}
Rockafellar, R.T.: Monotone operators and the proximal point algorithm. SIAM
journal on control and optimization \textbf{14}(5), 877--898 (1976)

\bibitem{chen2018efficient}
Chen, C. C., Yang, C. L., \& Cheng, H. Y. (2018). Efficient and robust parallel dnn training through model parallelism on multi-gpu platform. arXiv preprint arXiv:1809.02839.

\bibitem{masuyama2018modal}
Masuyama, Y., Kusano, T., Yatabe, K., \& Oikawa, Y. (2018, April). Modal decomposition of musical instrument sound via alternating direction method of multipliers. In {\it 2018 IEEE International Conference on Acoustics, Speech and Signal Processing (ICASSP)} (pp. 631-635). IEEE.

\bibitem{sun2018iteratively}
Sun, T., Jiang, H., Cheng, L., Zhu, W.: Iteratively linearized reweighted alter-
nating direction method of multipliers for a class of nonconvex problems. IEEE Transactions on Signal Processing \textbf{66}(20), 5380--5391 (2018)

\bibitem{sutskever2013importance}
Sutskever, I., Martens, J., Dahl, G., Hinton, G.: On the importance of initialization and momentum in deep learning. In: International conference on machine learning. pp. 1139--1147 (2013)

\bibitem{taylor2016training}
Taylor, G., Burmeister, R., Xu, Z., Singh, B., Patel, A., Goldstein, T.: Training neural networks without gradients: A scalable admm approach. In: International conference on machine learning. pp. 2722--2731 (2016)

\bibitem{tieleman2012lecture}
Tieleman, T., Hinton, G.: Lecture 6.5-rmsprop, coursera: Neural networks for machine learning. University of Toronto, Technical Report (2012)

\bibitem{wang2019admm}
Wang, J., Yu, F., Chen, X., Zhao, L.: Admm for efficient deep learning with global convergence. In: Proceedings of the 25th ACM SIGKDD International Conference on Knowledge Discovery \& Data Mining. pp. 111--119 (2019)

\bibitem{wang2019multi}
Wang, J., Zhao, L., Wu, L.: Multi-convex inequality-constrained alternating direction method of multipliers. arXiv preprint arXiv:1902.10882 (2019)

\bibitem{ouyang2013stochastic}
Ouyang, H., He, N., Tran, L., \& Gray, A. (2013, February). Stochastic alternating direction method of multipliers. In {\it International Conference on Machine Learning} (pp. 80-88).

\bibitem{zhong2014fast}
Zhong, W., Kwok, J.: Fast stochastic alternating direction method of multipliers. In: International Conference on Machine Learning. pp. 46--54 (2014)

\bibitem{Saeed1} 
Saeed G., Guanghui L.: Stochastic first- and zeroth-order methods for nonconvex stochastic programming. \textit{SIAM Journal on Optimization}, 23(4):2341-2368, 2013a. doi: 10.1137/ 120880811.

\bibitem{Saeed2}
Saeed G., Guanghui L., Hongchao Z. :Mini-batch stochastic approximation methods
for nonconvex stochastic composite optimization. \textit{Mathematical Programming}, 155(1-2):267-305,
2014.

\bibitem{zou2019sufficient}
Zou, F., Shen, L., Jie, Z., Zhang, W., Liu, W.: A sufficient condition for convergences of adam and rmsprop. In: Proceedings of the IEEE Conference on Computer Vision and Pattern Recognition. pp. 11127--11135 (2019)

\bibitem{chang2016asynchronous}
Chang, T. H., Hong, M., Liao, W. C., \& Wang, X. (2016). Asynchronous distributed ADMM for large-scale optimization-Part I: Algorithm and convergence analysis. IEEE Transactions on Signal Processing, 64(12), 3118--3130.

\bibitem{chang2016asynchronous2}
Chang, T. H., Hong, M., Liao, W. C., \& Wang, X. (2016). Asynchronous distributed ADMM for large-scale optimization-Part I: Algorithm and convergence analysis. IEEE Transactions on Signal Processing, 64(12), 3118-3130.

\bibitem{wei2013}
Wei, E., \& Ozdaglar, A. (2013, December). On the o (1= k) convergence of asynchronous distributed alternating direction method of multipliers. In {\it 2013 IEEE Global Conference on Signal and Information Processing} (pp. 551-554). IEEE.

\bibitem{jiang2019}
Jiang, S., Lei, Y., Wang, S., \& Wang, D. (2019, August). An Asynchronous ADMM Algorithm for Distributed Optimization with Dynamic Scheduling Strategy. In 2019 IEEE 21st International Conference on High Performance Computing and Communications; IEEE 17th International Conference on Smart City; IEEE 5th International Conference on Data Science and Systems (HPCC/SmartCity/DSS) (pp. 1-8). IEEE.

\bibitem{dodds2006popular}
Dodds, K. (2006). Popular geopolitics and audience dispositions: James Bond and the internet movie database (IMDb). {\it Transactions of the Institute of British Geographers}, 31(2), 116-130.

\bibitem{li2019distributed}
Li, Y., Wang, X., Fang, W., Xue, F., Jin, H., Zhang, Y., \& Li, X. (2019). A distributed ADMM approach for collaborative regression learning in edge computing. Comput. Mater. Contin, 59, 493-508.

\bibitem{chen2018fully}
Chen, G., \& Li, J. (2018). A fully distributed ADMM-based dispatch approach for virtual power plant problems. Applied Mathematical Modelling, 58, 300-312.

\bibitem{hosseini2014online}
Hosseini, S., Chapman, A., \& Mesbahi, M. (2014). Online distributed ADMM on networks. arXiv preprint arXiv:1412.7116.

\bibitem{sun2018alternating}
Sun, T., Yin, P., Cheng, L., \& Jiang, H. (2018). Alternating direction method of multipliers with difference of convex functions. Advances in Computational Mathematics, 44(3), 723-744.

\bibitem{guan2018an}
Guan, L., Qiao, L., Li, D., Sun, T., Ge, K., \& Lu, X. (2018, November). An efficient ADMM-based algorithm to nonconvex penalized support vector machines. In 2018 IEEE International Conference on Data Mining Workshops (ICDMW) (pp. 1209-1216). IEEE.











\end{thebibliography}
\end{document}